\documentclass[pdflatex]{sn-jnl}

\usepackage{graphicx}%
\usepackage{multirow}%
\usepackage{amsmath,amssymb,amsfonts}%
\usepackage{amsthm}%
\usepackage{mathrsfs}%
\usepackage[title]{appendix}%
\usepackage{xcolor}%
\usepackage{textcomp}%
\usepackage{manyfoot}%
\usepackage{booktabs}%
\usepackage{algorithm}%
\usepackage{algorithmicx}%
\usepackage{algpseudocode}%
\usepackage{listings}%
\usepackage{mathtools}
\usepackage{tabularx}
\usepackage[most]{tcolorbox}
\usepackage{letltxmacro}
\usepackage{xstring}
\usepackage{environ}

\theoremstyle{thmstyleone}%
\theoremstyle{thmstyletwo}%

\theoremstyle{thmstylethree}%

\usepackage[capitalize,noabbrev]{cleveref}

\definecolor{wikiblue}{HTML}{1195DB}
\definecolor{Grokipediaorange}{HTML}{F4A324} 
\definecolor{sciencepediagreen}{HTML}{27B57A}
\definecolor{baselinecolor}{HTML}{DC5151}

\newtcolorbox{wikipediaarticle}[1][Wikipedia Article]{
    colback=wikiblue!10!white,
    colframe=wikiblue,
    fonttitle=\bfseries,
    breakable,
    title={#1}
}

\newtcolorbox{sciencepediaarticle}[1][Sciencepedia Article]{
    colback=sciencepediagreen!10!white,
    colframe=sciencepediagreen,
    fonttitle=\bfseries,
    breakable,
    title={#1}
}

\newtcolorbox{llmbaseline}[1][Baseline LLM Article]{
    colback=baselinecolor!10!white,
    colframe=baselinecolor,
    fonttitle=\bfseries,
    breakable,
    title={#1}
}

\newtcolorbox{grokipediaarticle}[1][Grokipedia Article]{
    colback=Grokipediaorange!10!white,
    colframe=Grokipediaorange,
    fonttitle=\bfseries,
    breakable,
    title={#1}
}

\begin{document}

\title[Article Title]{Inverse Knowledge Search over Verifiable Reasoning: Synthesizing a Scientific Encyclopedia from a Long Chains-of-Thought Knowledge Base}


\author[1,2]{\fnm{Yu} \sur{Li}}
\equalcont{These authors contributed equally to this work.}
\author[3]{\fnm{Yuan} \sur{Huang}}
\equalcont{These authors contributed equally to this work.}
\author[4]{\fnm{Tao} \sur{Wang}}
\author[3,2]{\fnm{Caiyu} \sur{Fan}}
\author[2]{\fnm{Xiansheng} \sur{Cai}}
\author[6]{\fnm{Sihan} \sur{Hu}}
\author[3]{\fnm{Xinzijian} \sur{Liu}}
\author[5]{\fnm{Cheng} \sur{Shi}}
\author[3]{\fnm{Mingjun} \sur{Xu}}
\author[3]{\fnm{Zhen} \sur{Wang}}
\author[3]{\fnm{Yan} \sur{Wang}}
\author[7]{\fnm{Xiangqi} \sur{Jin}}
\author[8]{\fnm{Tianhan} \sur{Zhang}}
\author[7]{\fnm{Linfeng} \sur{Zhang}}
\author[4]{\fnm{Lei} \sur{Wang}}
\author[6,9]{\fnm{Youjin} \sur{Deng}}
\author[2,10,6]{\fnm{Pan} \sur{Zhang}}
\author[3]{\fnm{Weijie} \sur{Sun}}
\author[12]{\fnm{Xinyu} \sur{Li}}
\author[11,12,13]{\fnm{Weinan} \sur{E}}
\author*[3,12]{\fnm{Linfeng} \sur{Zhang}}\email{zhanglf@dp.tech}
\author*[1,2]{\fnm{Zhiyuan} \sur{Yao}}\email{yaozy@lzu.edu.cn}
\author*[2]{\fnm{Kun} \sur{Chen}}\email{chenkun@itp.ac.cn}

\affil[1]{\orgdiv{Lanzhou Center for Theoretical Physics, Key Laboratory of Theoretical Physics of Gansu Province, Key Laboratory of Quantum Theory and Applications of MoE, Gansu Provincial Research Center for Basic Disciplines of Quantum Physics}, \orgname{Lanzhou University}, \orgaddress{\city{Lanzhou}, \postcode{730000}, \country{China}}}

\affil[2]{\orgname{Institute of Theoretical Physics, Chinese Academy of Sciences}, \orgaddress{\city{Beijing}, \postcode{100190}, \country{China}}}

\affil[3]{\orgname{DP Technology}, \orgaddress{\city{Beijing}, \postcode{100080}, \country{China}}}

\affil[4]{\orgname{Institute of Physics, Chinese Academy of Sciences}, \orgaddress{\city{Beijing}, \postcode{100190}, \country{China}}}

\affil[5]{\orgname{}{D\'epartement d'Informatique, \'Ecole normale sup\'erieure}, \orgaddress{\city{Paris}, \postcode{75230}, \country{France}}}

\affil[6]{\orgdiv{Hefei National Laboratory}, \orgname{University of Science and Technology of China}, \orgaddress{\city{Hefei}, \postcode{230026}, \country{China}}}

\affil[7]{\orgdiv{School of Artificial Intelligence}, \orgname{Shanghai Jiao Tong University}, \orgaddress{\city{Shanghai}, \postcode{200240}, \country{China}}}

\affil[8]{\orgdiv{School of Astronautics}, \orgname{Beihang University}, \orgaddress{\city{Beijing}, \postcode{100191}, \country{China}}}

\affil[9]{\orgdiv{Hefei National Laboratory for Physical Sciences at the Microscale and Department of Modern Physics}, \orgname{University of Science and Technology of China}, \orgaddress{\city{Hefei}, \postcode{230026}, \country{China}}}

\affil[10]{\orgdiv{School of Fundamental Physics and Mathematical Sciences, Hangzhou Institute for Advanced Study}, \orgname{University of Chinese Academy of Sciences}, \orgaddress{\city{Hangzhou}, \postcode{310024}, \country{China}}}

\affil[11]{\orgdiv{School of Mathematical Sciences}, \orgname{Peking University}, \orgaddress{\city{Beijing}, \postcode{100871}, \country{China}}}

\affil[12]{\orgname{AI for Science Institute}, \orgaddress{\city{Beijing}, \postcode{100080}, \country{China}}}

\affil[13]{\orgdiv{Center for Machine Learning Research}, \orgname{Peking University}, \orgaddress{\city{Beijing}, \postcode{100871}, \country{China}}}


\abstract{
Most scientific materials compress reasoning, presenting conclusions while omitting the derivational chains that justify them. This compression hinders verification by lacking explicit, step-wise justifications and inhibits cross-domain links by collapsing the very pathways that establish the logical and causal connections between concepts. We introduce a scalable framework that decompresses scientific reasoning, constructing a verifiable Long Chain-of-Thought (LCoT) knowledge base and projecting it into an emergent encyclopedia, SciencePedia. Our pipeline operationalizes an endpoint-driven, reductionist strategy: a Socratic agent, guided by a curriculum of around 200 courses, generates approximately 3 million first-principles questions. To ensure high fidelity, multiple independent solver models generate LCoTs, which are then rigorously filtered by prompt sanitization and cross-model answer consensus, retaining only those with verifiable endpoints. This verified corpus powers the Brainstorm Search Engine, which performs inverse knowledge search---retrieving diverse, first-principles derivations that culminate in a target concept. This engine, in turn, feeds the Plato synthesizer, which narrates these verified chains into coherent articles. The initial SciencePedia comprises approximately 200,000 fine-grained entries spanning mathematics, physics, chemistry, biology, engineering, and computation. In evaluations across six disciplines, Plato-synthesized articles (conditioned on retrieved LCoTs) exhibit substantially higher knowledge-point density and significantly lower factual error rates than an equally-prompted baseline without retrieval (as judged by an external LLM). Built on this verifiable LCoT knowledge base, this reasoning-centric approach enables trustworthy, cross-domain scientific synthesis at scale and establishes the foundation for an ever-expanding encyclopedia.
}




\maketitle

\section{Introduction}
\label{submission}

Wikipedia stands as a monumental achievement of the information age, a testament to the power of human curation. Yet, for all its breadth, it exhibits systemic limitations: quality is inconsistent across languages~\cite{wiki_language1,wiki_language2}, disciplinary silos are difficult to breach~\cite{wiki_topic1}, and verifying complex claims is challenging. We argue that these are not separate flaws, but symptoms of a single, fundamental cause inherent to the human curation paradigm: the radical compression of reasoning. Due to the immense cost of human time and effort, our scientific corpus---from textbooks to wikis---is economically forced to prioritize conclusions over the reasoning that produces them.

This compression of reasoning has profound consequences. It omits the explicit chains of logic, derivation, and synthesis that form the very structure of knowledge. We argue that this vast, unrecorded network of reasoning is the \textbf{``dark matter" of human knowledge}. In cosmology, dark matter is the invisible mass whose gravitational effects give structure to the visible universe~\cite{darkmatter}; similarly, this intellectual dark matter is the latent, connective scaffolding that underpins and shapes the explicit facts we record. Its omission leads to two critical failures in our current knowledge systems. First, without this explicit scaffolding, knowledge becomes difficult to navigate and verify; trust must be placed in authority rather than in a transparent, auditable thought process. Second, and more critically, by compressing away the derivational pathways, we sever the intrinsic connections within and between fields, losing the subtle, cross-disciplinary links that drive innovation.

Reversing this compression---externalizing this ``dark matter" of reasoning---would require an engine capable of generating and validating knowledge at a scale far beyond human capacity. The advent of Large Language Models (LLMs) offers the first plausible candidate for such an engine~\cite{LLM}. Yet this raises a fundamental challenge: can a tool trained on a compressed map recreate the full territory? A naive approach---simply tasking an LLM to distill knowledge or write an encyclopedia---is doomed to fail. Because the LLM is designed to reproduce the patterns in its training data, it will faithfully replicate the same compressed, conclusion-oriented format. It will inherit the ``dark matter blindness" of the human internet corpus. Moreover, LLMs are prone to hallucinations---generating plausible but factually incorrect information~\cite{hallucination1,hallucination2,hallucination3,ji2023survey,hallucination5}---which poses severe challenges for knowledge verification and reliability.

Therefore, to establish a scientific knowledge base that transcends the limitations of both compressed human-curated corpora and fallible LLM generation, our solution is a deliberate, two-step process: first, the systematic construction of a massive, verifiable, and deeply interconnected Long Chain-of-Thought (LCoT) knowledge base rooted in first-principles derivations~\cite{cot}; and second, its projection into a human-explorable encyclopedia. 
 
 In stark contrast to human internet corpora, our framework roots knowledge in first principles by operationalizing a reductionist strategy. It begins by defining a comprehensive set of knowledge points from a curriculum of approximately $200$ courses. From these endpoints, a Socratic method ~\cite{socratic1,socratic2,socratic3} is employed to automatically generate a corpus of around three million first-principles-based questions. Each question is then processed by multiple, distinct LLMs to generate and cross-validate a comprehensive derivational path using LCoT. This methodology yields key advantages: the detailed reasoning chains not only establish verifiable connections between concepts---thus externalizing the ``dark matter" of knowledge---but also enable the exhaustive validation of the entire logical structure, all within an inherently scalable, LLM-driven architecture.

Our second contribution is the projection of this uncompressed knowledge graph into a new kind of scientific resource. We develop a search paradigm that operates directly on the reasoning chains, facilitating a form of ``inverse search". This mechanism, which we term the \textbf{Brainstorm Search Engine}, retrieves the diverse, non-trivial, and cross-disciplinary pathways that lead to a given concept. It forms the backbone for our second contribution: \textbf{SciencePedia}\footnote{SciencePedia Homepage: \href{sciencepedia.bohrium.com}{sciencepedia.bohrium.com}}. This encyclopedia is not written in a traditional sense but is instead emergent---for a given knowledge point, its article is generated through the automated synthesis of all relevant derivational chains retrieved by the Brainstorm Search Engine. The initial version already comprises approximately 200,000 entries spanning mathematics, physics, chemistry, biology, engineering, and computational science. Critically, the framework is built not just for finding facts, but for discovering the profound connections that link all scientific domains.

It is important to note that our LLM-driven automated pipeline is primarily designed to solve the ``cold start" problem inherent in building such a large-scale, deeply interconnected scientific encyclopedia. By generating a solid, verifiable foundation from first principles, we provide a robust starting point for future development. We envision that this automated generation can be combined with traditional community-driven efforts in the future, leveraging the expertise of human specialists to further refine knowledge coverage, add new insights, and correct potential errors, thereby creating an evolving and self-correcting knowledge ecosystem.

The remainder of this paper is structured as follows. Section \ref{sec:knowledgebase} details the methodology behind our Socrates agent, which systematically constructs the LCoT knowledge base through a Socratic questioning process. Section \ref{sec:search_engine} introduces the Brainstorm Search Engine, a novel search mechanism for discovering cross-domain connections within this knowledge base, and discusses its application in building the Plato agent, a creative writer designed for high-divergence and low-hallucination synthesis. Finally, Section \ref{sec:sciencepedia} elaborates on how these components are integrated to construct SciencePedia.

\section{Why LCoTs Form a New and Reliable Corpus}
\label{sec:lcot_corpus}

We posit that a corpus of LCoTs generated by modern LLMs constitutes a novel data distribution, fundamentally distinct from the internet corpus upon which they were pre-trained. This distinction, and the reliability of our resulting knowledge base, rests on two pillars.

\subsection{Novelty: A New Statistical Distribution of Reasoning}
Standard pre-training aligns a Base Model (henceforth $p_{\text{Base}}$) with the distribution of the human internet corpus ($p_{\text{Internet}}$). This corpus is fundamentally ``compressed": it is dense with ``System 1" facts and conclusions but exceedingly sparse in the ``System 2" explicit, step-by-step derivations that justify them. Let $Q$ be an input question and $\text{LCoT}$ be a specific Long Chain-of-Thought derivation. The probability of finding such a chain in the training data is near zero. Consequently, the Base Model, which mimics this distribution, has almost no capacity to generate them:
\begin{equation}
p_{\text{Base}}(\text{LCoT} | Q) \approx p_{\text{Internet}}(\text{LCoT} | Q) \approx 0
\label{eq:p_base}
\end{equation}

A fundamental shift occurs during post-training. Reinforcement Learning from Verifiable Rewards (RLVR) is a key post-training paradigm that optimizes a language model not to mimic surface text, but to produce reasoning trajectories whose endpoints can be mechanically checked (e.g., passing a unit test or matching a known solution)~\cite{deepseek-r1, team2025kimi}. This unlocks the new, emergent capability for long-form reasoning~\cite{cai2025learning, hu2025llms}. The RLVR-trained model (henceforth $p_{\text{LLM}}$) is now capable of generating extensive, multi-step LCoTs in response to a prompt $Q$. This creates a vast statistical deviation between the two generative distributions:
\begin{equation}
p_{\text{LLM}}(\text{LCoT} | Q) \gg p_{\text{Base}}(\text{LCoT} | Q) \approx 0
\label{eq:p_llm}
\end{equation}
This statistical deviation (Eq. \ref{eq:p_llm}) is the novelty of our corpus. 

\subsection{Reliability: Harnessing Inherent Causal and Logical Consistency}\label{sec:reliability}

The LCoTs from $p_{\text{LLM}}$ are not just statistically novel; they are qualitatively valuable. The RLVR process, by optimizing for verifiable outcomes, produces reasoning trajectories that inherently capture the latent long-range causal and logical links of science---the very ``dark matter" absent from the internet corpus.

This emergent $p_{\text{LLM}}$ distribution, while a high-fidelity source of reasoning, is not perfect. It still contains residual stochastic errors (``hallucinations"). However, a key objective property of these LCoTs is that they are inherently verifiable. Because these reasoning trajectories are trained to terminate at mechanically checkable endpoints (e.g., a numerical answer, a symbolic formula), their validity can be externally assessed.

This inherent verifiability is what makes their residual hallucinations controllable. It is statistically improbable for multiple, distinct models to independently generate different flawed reasoning paths that all converge, by coincidence, on the same correct, verifiable answer. Therefore, a consensus on the final answer, such as one obtained via Cross-Model Answer Validation, serves as a powerful and objective filter for causal and logical consistency. This property allows for the isolation of a corpus that is not just statistically novel but also possesses an extremely high fidelity to the true structures of science.

\section{LCoT Scientific Knowledge Base}
\label{sec:knowledgebase}

To address the challenge of building a large-scale, verifiable scientific knowledge base, we must systematically extract, validate, and structure the vast, yet often implicit and unreliable, knowledge latent within LLM \cite{petroni2019language,pan2023logic}. Our primary objective is to generate a repository that is comprehensive, deeply interconnected, and grounded in first principles. This requires a novel framework that moves beyond standard distillation techniques to explicitly externalize and verify the ``dark matter" of scientific reasoning.

The fundamental unit of knowledge within our architecture is the Long Chain-of-Thought Question-Answer (LCoT-QA) pair. This structure is paramount for three reasons. Firstly, it naturally mirrors the process of scientific inquiry, where a well-posed question serves as the starting point for deep exploration. Secondly, the QA format provides a clear framework for verification: a specific question often has a concrete, verifiable conclusion (e.g., a numerical answer or a symbolic derivation), allowing for the validation of the entire reasoning chain that produces it. Finally, this structure renders the vast knowledge base navigable, with questions acting as semantic entry points into the complex web of derivations.

\begin{figure}[h!]
    \centering
    \includegraphics[width=0.55\linewidth]{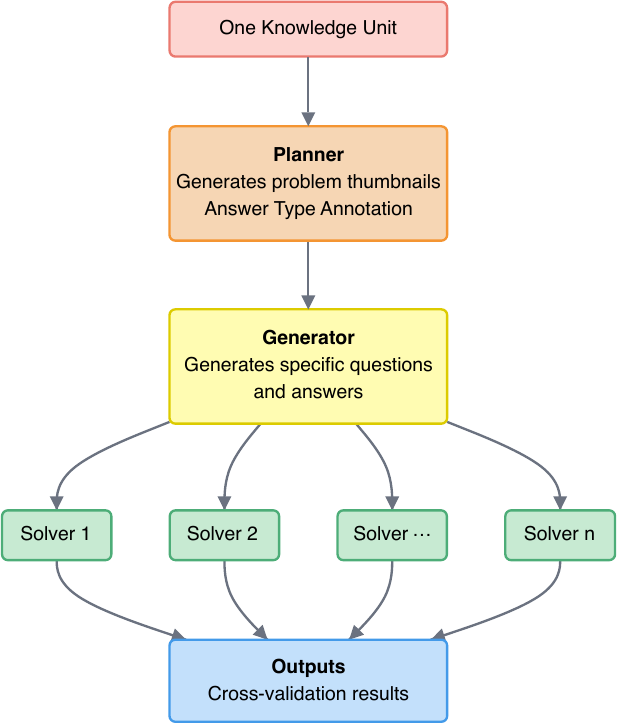}
	\caption{The three-stage process of problem generation and cross-validation. First, a \textbf{Planner} agent generates high-level `thumbnails' for problems based on a knowledge unit. Second, a \textbf{Generator} agent expands these thumbnails into specific questions with verifiable answers. Finally, the question is posed to multiple independent \textbf{Solver} agents (distinct LLMs), and their answers are cross-validated to ensure the correctness and reliability of the generated content.}
	\label{fig:planner_generator_solver}
\end{figure}

To generate these LCoT-QA pairs at scale, we developed the Socrates agent, a systematic framework whose workflow is illustrated in Fig.~\ref{fig:planner_generator_solver}. This bespoke framework was necessary, as standard methods like knowledge distillation~\cite{hinton2015distilling,gou2021knowledge} are ill-suited for this task. While effective for their primary application (e.g., training data generation), these methods are typically seeded by existing texts and optimized for summarization or rewriting, producing fragmented knowledge, lacking broad, multi-level disciplinary coverage, and failing to capture the long derivational pathways from first principles. It captures the ``what" but misses the crucial ``why", leaving the logical connections that form the bulk of scientific knowledge implicit\cite{huang2022towards,sejnowski2023large}.

\subsection{An Endpoint-Driven, Reductionist Strategy}

To overcome the limitations described above, the Socrates agent operationalizes an inverse, endpoint-driven strategy inspired by the scientific principle of reductionism. Reductionism posits that a complex system can be understood by analyzing its constituent parts \cite{weinberg1987newtonianism,gross2005discovery}. Instead of providing a model with a set of axioms and asking it to reason ``forward"---a process whose completeness is difficult to guarantee---we start with a high-level knowledge point (an ``endpoint") and task the model with deriving it from more fundamental principles \cite{russell1995artificial}. This methodology has two critical advantages.

First, by sampling the endpoints of reasoning chains, we can systematically ensure completeness. While curating a complete set of scientific axioms is nearly impossible, compiling a representative set of key concepts and theorems from established curricula is tractable. By ensuring broad coverage of these endpoints, we induce comprehensive coverage of the underlying principles required for their derivation.

Second, to generate information-rich chains, we prompt the model to derive the same endpoint from multiple, distinct levels of abstraction (e.g., from high-school, undergraduate, and graduate-level principles). This forces the model to articulate the connections between different layers of scientific understanding.

Taken together, this systematic deconstruction of knowledge---moving from a complex conclusion back to its fundamental premises through targeted, layered questioning---constitutes a modern, scalable implementation of the Socratic method.

\subsection{Scalable Implementation via Curriculum Scaffolding}
To define a comprehensive set of endpoints at scale, we developed a systematic, curriculum-based scaffolding process. We began by manually curating a broad academic curriculum of approximately 200 undergraduate- and graduate-level courses across major scientific disciplines. For each course, we enumerated approximately 200 core topics.

For each topic, Socrates automatically generates a diverse set of around 100 prompts, which fall into two main categories:
\begin{enumerate}
\item \textbf{Reductionist Prompts (``What and Why"):} These ask the model to explain a concept or derive a result from specified first principles. For example, ``Explain the physical significance of the two constants of motion, energy and angular momentum, in determining the trajectory of a body under a central force". In a similar spirit, one may also encounter more advanced derivation-type questions such as ``Derive the semiclassical equations of motion for a Bloch wavepacket in an external electric field, $\dot{\vec r}=\nabla_{\vec k} \varepsilon_n(\vec k)-\dot{\vec k}\times\mathcal F_n(\vec k)$ and $\hbar\dot{\vec k}=-e\vec E$, starting from a Lagrangian with Berry connection". Both types of questions require reasoning from first principles---whether classical mechanics or quantum mechanics---to reconstruct a result or interpret its physical meaning step by step.
\item \textbf{Application Prompts (``How"):} These ground theory in practical contexts, asking how a principle is applied or a phenomenon is utilized. For example, ``Use a simple pendulum of known length to determine the acceleration due to gravity $g$ on a hypothetical planet, given its measured period of oscillation." or ``Explain how a Cherenkov detector can be used to distinguish between two particles of different mass but the same high momentum.". Such questions connect theoretical understanding to experimental or technological situations, requiring the application of physical laws to analyze measurable quantities, design methods, or interpret real-world observations.
\end{enumerate}

This three-level generation process---from courses to topics to prompts---allows Socrates to scalably produce a massive and diverse set of high-quality LCoT queries, forming the raw material for our knowledge base.

\subsection{Knowledge Verification: A Multi-Faceted Protocol}

Following generation, the raw corpus of LCoT-QA pairs is subjected to a rigorous verification protocol to ensure its logical consistency and factual accuracy. To address the challenge of LLM hallucination, this protocol establishes strong guards at both ends of the reasoning chain.

It begins with Prompt Sanitization. Before any reasoning is attempted, the agent first screens the initial question set it generated. Using a distinct LLM, it checks for scientific inaccuracies, flawed assumptions, or unreasonable values in the prompts themselves \cite{madaan2023self}. This process filters out approximately $5\%$ of automatically generated problems, preventing the model from reasoning based on a faulty premise.

The process is further strengthened by a \textbf{Verifiable Endpoint Design}. The prompt generation strategy is intentionally biased towards questions with objectively verifiable answers, such as those requiring symbolic/numerical calculations, coding solutions, or multiple-choice selections. This transforms the abstract problem of verifying a complex reasoning chain into the more tractable problem of validating a concrete answer.

Finally, the agent performs \textbf{Cross-Model Answer Validation}. To validate the final answer and, by extension, the reasoning process, each prompt is processed by at least two distinct LLMs from different providers. If the models produce divergent final answers, the entire QA pair is flagged as unreliable and discarded. The necessity of this check is underscored by our findings: in a sample of physics questions, the success rate of LLMs dropped from $\sim 70\%$ for undergraduate problems to $\sim 50\%$ for graduate topics, highlighting the need for this rigorous validation step.

This dual verification---sanitizing the starting points (prompts) and validating the endpoints (answers)---significantly enhances the reliability of the entire knowledge base.

\begin{figure*}[htbp]
    \centering
    \includegraphics[width=0.85\linewidth]{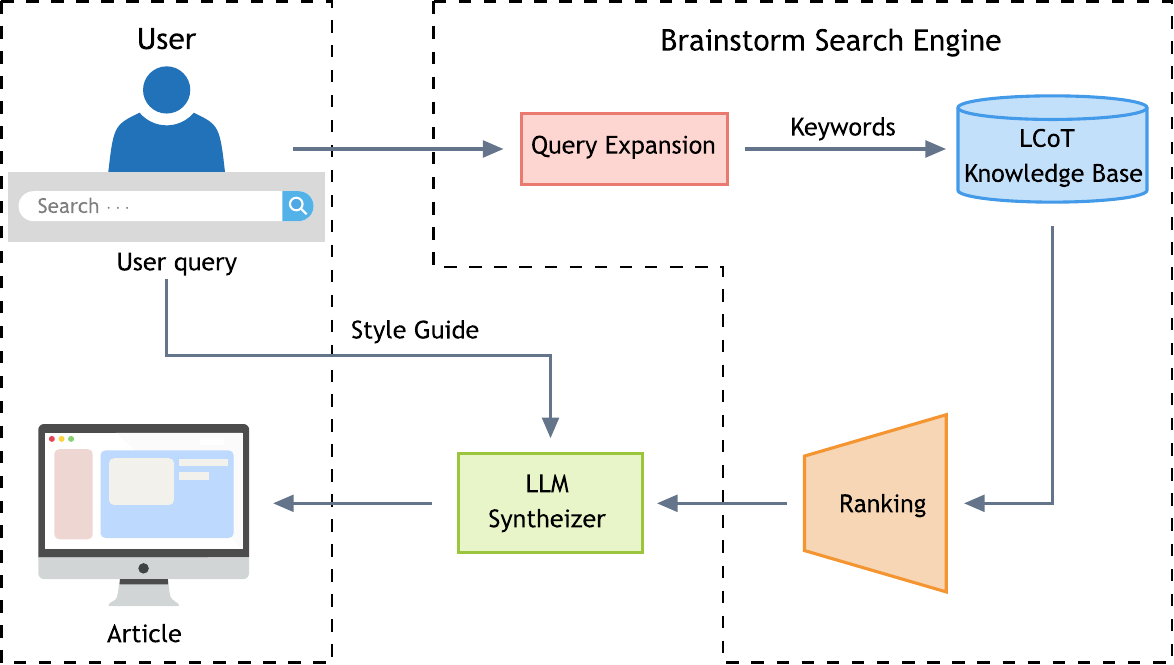}
    \caption{
        \textbf{The Brainstorm Search Engine and Plato Agent Architecture.} A user initiates a query (e.g., a target knowledge point) for the Brainstorm Search Engine. The \textbf{Query Expansion} module processes this input into keywords to retrieve relevant Long Chain-of-Thought (LCoT) derivations from the \textbf{LCoT Knowledge Base}. These derivations, representing the ``dark matter" of scientific reasoning, are then ranked based on their relevance and cross-disciplinary significance. The ranked LCoT derivations serve as a verifiable ``reasoning scaffold" for the \textbf{LLM Synthesizer (Plato Agent)}. Guided by the user's initial query and an optional \textbf{Style Guide}, the Plato Agent synthesizes these verified derivations into a coherent and pedagogically clear article. This architecture enables ``inverse knowledge search", transforming search into a discovery process that reveals the provenance and interconnections of scientific concepts, while mitigating hallucination through grounding in the LCoT knowledge base.
    }
    \label{fig:plato_based_brainstorm_search_engine}
\end{figure*}

\section{The Brainstorm Search Engine for Knowledge Discovery}
\label{sec:search_engine}

Once the LCoT Knowledge Base is constructed, it provides the foundation for a novel tool for knowledge synthesis and discovery: the Brainstorm Search Engine. This engine is designed to overcome the fundamental limitations of both traditional search engines and modern AI agents.

Traditional search engines, such as Google or Bing, are optimized to index the human internet---a corpus dominated by the conclusions of reasoning, not the reasoning process itself \cite{brin1998anatomy}. Due to the radical compression of reasoning inherent in wikis, articles, and textbooks, the derivational pathways, or ``dark matter", are largely invisible to their crawlers. A user can find what a concept is, while uncovering the rich, cross-disciplinary context of how it is derived or why it is important remains a significant challenge.

Modern deep research AI agents, which rely on this very corpus for information retrieval and synthesis, consequently inherit its fundamental limitations \cite{bender2021dangers}. This inheritance manifests in two critical ways. First, when synthesizing content from sources dense with facts but devoid of explicit reasoning, the agents operate with a superficial understanding of the material, leading to a high risk of factual hallucination \cite{hallucination1,hallucination2,hallucination3,ji2023survey,hallucination5}. Second, because the crucial cross-disciplinary links---the very ``dark matter" of knowledge---are absent from their source data, these agents are inherently challenged in discovering the non-trivial connections that drive scientific insight \cite{yao2023tree}.

The Brainstorm Search Engine directly addresses these challenges by operating on the LCoT Knowledge Base, a dataset where the reasoning process is the primary content.

\subsection{Inverse Knowledge Search: From Concept to Provenance}

The engine's core mechanism is a paradigm we term ``inverse knowledge search". Instead of querying for a definition or a fact (the endpoint of reasoning), a user provides a target concept, and the engine retrieves the diverse collection of LCoT derivational chains that feature this concept within their reasoning process. Because every chain in our knowledge base is grounded in first principles and rigorously verified, the retrieved pathways exhibit high scientific integrity \cite{liang2024survey}.

This process allows users to directly access the ``dark matter" of human knowledge: the rich derivational pathways and cross-disciplinary links omitted from traditional texts \cite{swanson1986undiscovered}. It facilitates a mode of exploration that mirrors scientific discovery, revealing a concept's context, prerequisites, and implications \cite{stokes2020deep}.

For example, a conventional search for ``Instanton" would likely return its technical definition. Our engine, in contrast, reveals the rich tapestry of its origins and applications by presenting the LCoT derivations that show its fundamental role as a descriptor of quantum tunneling \cite{belavin1975pseudoparticle,coleman1988aspects}, first conceptualized in simple systems like the double-well potential \cite{Polyakov_1977qc}; its profound implications in the Standard Model, explaining the structure of the QCD vacuum and the violation of baryon number \cite{hooft1976symmetry}; its application in cosmology, where gravitational instantons describe processes like Hawking radiation \cite{gibbons1977action}; and its surprising utility in pure mathematics, leading to breakthroughs in the understanding of 4-dimensional manifolds \cite{donaldson1983application}.

By exposing these non-trivial connections, the Brainstorm Search Engine transforms search from a simple lookup into an act of exploration and discovery.

\begin{figure}[h]
    \centering
    \includegraphics[width=\linewidth]{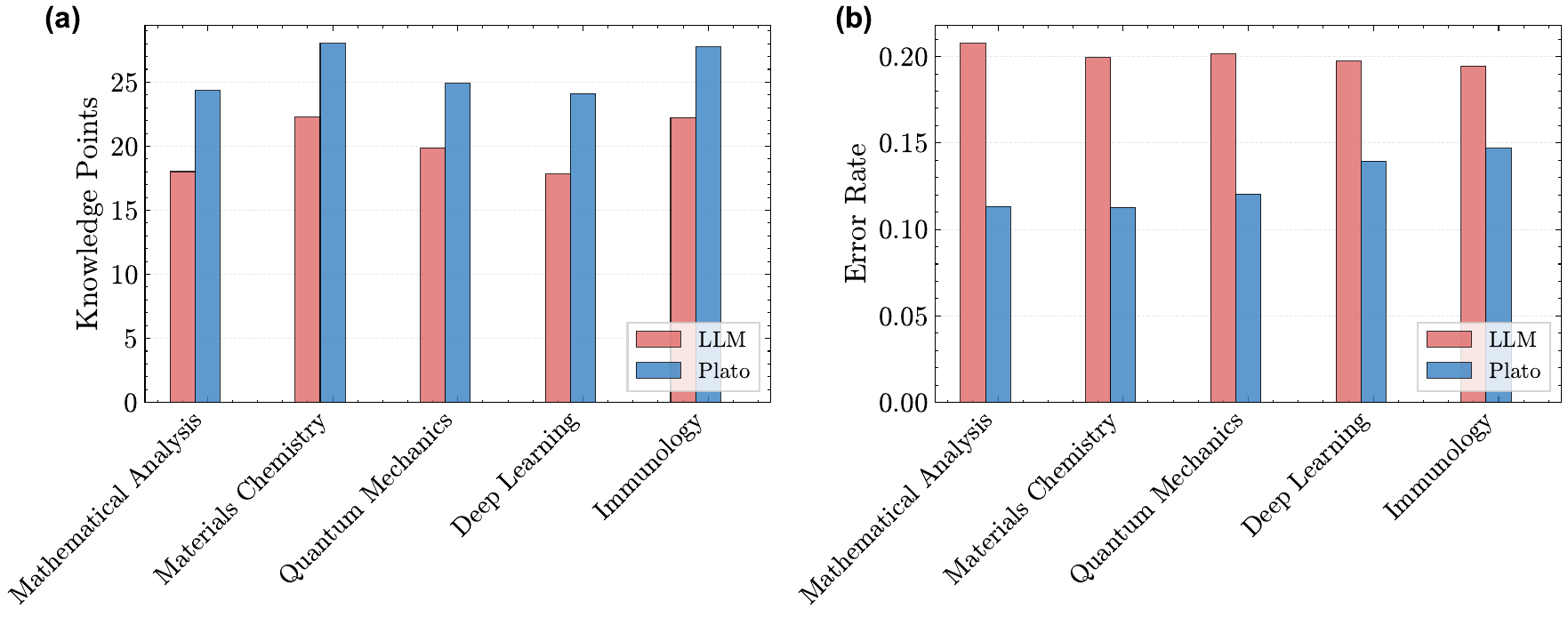}
    \label{fig:knowledge_points}
    \label{fig:error_rate}
    \caption{
        \textbf{(a) Comparison of Knowledge Point Density.} This figure compares the number of unique, learnable knowledge points contained in articles generated for the same set of topics. Both methods were tasked with generating articles using essentially identical prompts, with the only difference being whether access to verified LCoT context is provided. The \textbf{LLM} baseline received no retrieved LCoT context (an empty set), while \textbf{Plato} received the LCoT corpus from the Brainstorm Search Engine to synthesize its articles. Across all tested scientific domains, the Plato agent consistently produces articles with significantly greater knowledge density, demonstrating the superior depth and comprehensiveness of our synthesis approach.
        \textbf{(b) Comparison of Factual Error Rates.} This figure evaluates the factual reliability of articles generated using essentially identical prompts for both methods (as evaluated by GPT-5). The \textbf{LLM} baseline, which received no retrieved LCoT context (an empty set), exhibits a high error rate indicative of model hallucination. In contrast, the \textbf{Plato} agent, which received the retrieved LCoT corpus, grounds its synthesis in this pre-verified knowledge and achieves a significantly lower error rate across all domains. This highlights the effectiveness of our approach in producing highly reliable scientific content.
    }
\end{figure}

\subsection{The Plato Agent: High-Fidelity Synthesis}

This capacity for discovering novel, verified connections provides a direct solution to the problem of hallucination in AI-driven scientific writing. The Plato agent is a creative synthesizer built upon the Brainstorm Search Engine. Its task is not unconstrained generation but structured synthesis \cite{lewis2020retrieval}. 
The schematic architecture of the search-and-synthesis pipeline is depicted in Fig.~\ref{fig:plato_based_brainstorm_search_engine}.

The rich, cross-disciplinary ``reasoning scaffolds" retrieved by the search engine serve as a verifiable foundation that resolves the typical trade-off between creativity and factual accuracy. The creativity of any resulting synthesis is not the product of a model's stochastic generation but is inherited directly from the surprising and verified connections surfaced by the search. Simultaneously, this grounding mechanism dramatically reduces hallucination, as the synthesis is anchored to these explicit and pre-verified reasoning scaffolds \cite{shuster2021retrieval,rashkin2021increasing,ji2023survey}. 

An LLM's task within the Plato agent is thus shifted from pure generation to narration: weaving the interdisciplinary examples from the provided scaffold into a unified and pedagogically clear narrative. For instance, using the ``Instanton" scaffold, the agent constructs a story beginning with quantum tunneling and guiding the reader to its applications in QCD and cosmology. The LLM's role is focused on building narrative bridges between verified concepts.

This improvement is quantitatively validated by our evaluations. We compared Plato-synthesized articles (grounded in retrieved LCoT scaffolds) against a strong LLM baseline (generated from the same prompt, adjusted solely by the absence of retrieved context) for a scientific encyclopedia-style writing task across six scientific disciplines (See Appendix \ref{app:transmon} for an example). The results are clear: as shown in Fig.~\ref{fig:knowledge_points} , Plato consistently achieves a significantly higher knowledge-point density. Simultaneously, as shown in Fig.~\ref{fig:error_rate}, it achieves a substantially lower factual error rate---reducing hallucinations by approximately $50\%$---confirming the effectiveness of grounding synthesis in our explicit, pre-verified LCoT Knowledge Base.

Ultimately, the Brainstorm Search Engine acts as a discovery tool, surfacing novel connections that then serve as the verifiable backbone for the Plato agent's creative synthesis.

\begin{figure*}[h!]
    \centering
    \includegraphics[width=1.0\linewidth]{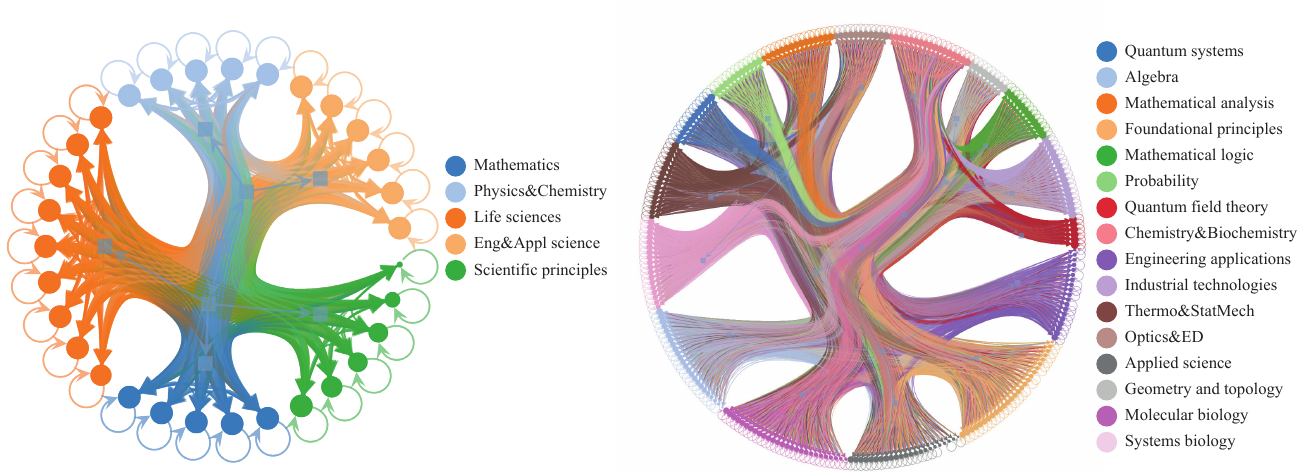}
    \caption{
\textbf{Hierarchical structure of the keyword graph.} We applied the modularity belief propagation algorithm~\cite{zhang2014scalable,shi2018weighted} to cluster the 120,226 keyword nodes. This process identified 7,454 base communities and yielded a hierarchical structure spanning 21 levels. To illustrate, nodes are progressively coarsened into cluster nodes at each level. The left panel shows the aggregation at level 3, and the right panel shows the aggregation at level 5. Figures were produced with the \texttt{graph-tool} package~\cite{peixoto2017graph}, and community titles were summarized by an LLM.
}
    \label{fig:clustering}
\end{figure*}

\section{SciencePedia: An Encyclopedia Emerged from Long Chains of Thought}
\label{sec:sciencepedia}

The verified LCoT knowledge base, constructed by the Socrates agent, serves as the foundation for an important application: the creation of \textbf{SciencePedia}, a comprehensive, cross-disciplinary STEM encyclopedia. The core hypothesis is that a sufficiently large and diverse collection of LCoT-QA pairs, each detailing a reasoning process from first principles, implicitly forms a dense network of knowledge. The connections are not pre-defined in a formal graph structure but emerge naturally from the content of the derivations themselves. When a concept appears in reasoning chains across different domains, an intrinsic, verifiable link between those domains is established. This section details the methodology for systematically transforming this vast repository of reasoning into a human-readable, deeply insightful, and highly reliable encyclopedia.

\subsection{The Automated Page Generation Workflow}

The process of generating an encyclopedia page is a deterministic workflow that leverages the Plato agent based on the Brainstorm Search Engine to translate raw LCoT-QA pairs into a structured narrative. The encyclopedia's structure originates from the curriculum defined during the Socrates agent's knowledge generation phase. This curriculum consists of approximately $200$ courses, each typically containing around $200$ core topics. For each topic, an LLM extracts about 10 representative keywords. After deduplication, this process yields a fine-grained list of approximately $200,000$ keywords, each seeding a unique encyclopedia page. This automatically forms a two-tiered structure of coarse-grained topics and fine-grained encyclopedia pages.

The generation workflow for each page begins with a \texttt{keyword} from this list (e.g., ``Instanton'') which seeds a search across the entire knowledge base, retrieving all LCoT-QA pairs where the keyword appears. An LLM then processes this collection, thematically organizing the pairs into two fundamental categories: those that address the concept's first principles (``What \& Why''), and those that demonstrate its use (``Application''). Finally, a specialized authoring LLM synthesizes this categorized material into a coherent article. The ``What \& Why'' pairs form a core section on \textbf{Principles and Mechanisms}, while the ``Application'' pairs are woven into a section on \textbf{Cross-Domain Applications}, naturally embedding cross-disciplinarity into the article's structure. In the future, new keywords can be systematically extracted from the content of existing encyclopedia pages, enabling a recursive expansion to cover even more granular knowledge points.

\subsection{Advantages over LLMs and Traditional Encyclopedias}

This methodology provides significant advantages over both querying a general-purpose LLM directly and relying on human-curated resources like Wikipedia. Compared to a direct LLM query, our approach ensures greater \textbf{depth}, as demonstrated by the higher knowledge-point density in our generated articles. The greater depth lies in our use of fine-grained, ``pre-digested" LCoT-QA pairs, which enable the model to internalize deeper conceptual linkages. This grounding in a pre-verified knowledge base also confers \textbf{high reliability}, drastically reducing the risk of factual hallucination~\cite{hallucination1,hallucination2,hallucination3,ji2023survey,hallucination5}. Moreover, because the search process systematically gathers every instance of a concept, the articles possess an \textbf{inherent cross-disciplinarity} that a single generative query might miss.

SciencePedia's automated approach also addresses fundamental limitations of human-curated wikis. It overcomes challenges of \textbf{scalability and language parity}~\cite{wiki_language1,wiki_language2}, as the pipeline can generate high-quality pages consistently across languages, unlike the volunteer-dependent model which often results in coverage disparities~\cite{wiki_topic1}. It also provides greater \textbf{explanatory depth}, moving beyond the ``checklist'' style of some articles to unfold the underlying reasoning process. Lastly, where a human-written page is limited by the author's breadth of knowledge, our search-driven method ensures a \textbf{systematic and comprehensive coverage of cross-domain applications}, providing a more complete view of a concept's role in science~\cite{cyberchiefs}.

\subsection{Writing Philosophy and Current Scope}

To maximize accessibility, the writing style of SciencePedia is inspired by Richard Feynman's ``Feynman Lectures on Physics"~\cite{feynman1964lectures}, focusing on advanced popular science that is intuitive and insightful while minimizing dense formulas and jargon. This pedagogical approach is intentionally distinct from the typically drier style of Wikipedia, though our framework is flexible enough to produce more technical articles by adjusting the authoring prompts.

It is important to note the current scope and limitations. SciencePedia currently covers major disciplines including mathematics, physics, chemistry, biology, engineering, and computational science. Since the knowledge base is primarily generated by LLMs, it contains a wealth of objective facts and logical reasoning but largely omits human-centric information, such as scientific history. Furthermore, its knowledge is bounded by the training data of the foundational models and thus lacks information on the latest scientific frontiers~\cite{knowledge_cutoff1}. In the future, we plan to address these gaps by applying the same LCoT-based digestion methodology to other corpora, such as textbooks and peer-reviewed articles. This will allow us to incorporate the rich history of scientific discovery, continuously update SciencePedia with cutting-edge research, and expand its coverage to other natural sciences like astronomy, geography, and economics, further enhancing its value as a truly comprehensive and dynamic resource.

\subsection{Hierarchical structures of the keyword graph}

A central hypothesis of this work is that by synthesizing articles from first-principles LCoTs, SciencePedia will naturally capture the deep, cross-disciplinary connections—the "dark matter"—that are omitted from traditional, compressed encyclopedias. To visualize and empirically validate this emergent interconnectivity, we performed a large-scale network analysis.First, we constructed a directed keyword graph. The encyclopedia entries serve as the nodes of this graph. For each encyclopedia page (representing a node $k_i$), we employed an LLM to parse its synthesized content and extract approximately ten most relevant keywords ($k_{j_1}, k_{j_2}, \dots$) that it references. A directed edge was then created from $k_i$ to each referenced node $k_j$. This process results in a large-scale graph that implicitly encodes the knowledge flow and conceptual dependencies within SciencePedia.

To make this large-scale structure visible, we perform graph clustering on this network. However, due to the graph's heterogeneity and the rich cross-domain interactions among scientific areas, conventional methods (e.g., $k$-means on embeddings or off-the-shelf agglomerative clustering) are either ill-suited to network community structure or computationally prohibitive at this scale.

We therefore adopt a statistical approach based on modularity belief propagation (MODBP)~\cite{zhang2014scalable,shi2018weighted}. 
Concretely, we first use MODBP to partition the graph into communities, and then recursively apply the same procedure to each induced subgraph. The recursion terminates once MODBP detects no further community structure, i.e., when all subgraphs are statistically indistinguishable from a random, structureless graph. 

This top-down procedure yields a tree-like hierarchy whose higher levels expose coarse scientific areas while lower levels refine into more specific topics, making cross-domain linkages explicit, as illustrated in Figure~\ref{fig:clustering}.

Crucially, this analysis does not merely reveal a set of siloed domains. We observe a high density of non-trivial connections between communities at various levels of the hierarchy. These inter-community links provide direct, empirical evidence for the emergent cross-disciplinarity our LCoT-based synthesis process captures, confirming that SciencePedia systematically surfaces the connections bridging disparate fields of study.

\section{Conclusion}

In this work, we addressed a fundamental limitation of our existing knowledge systems: the radical compression of reasoning. We argued that this compression obscures the derivational pathways---the ``dark matter" of knowledge---hindering verification, stifling cross-disciplinary discovery, and creating a fertile ground for AI hallucination~\cite{hallucination1,hallucination2,hallucination3,ji2023survey,hallucination5}. Our primary contribution is a comprehensive framework to systematically decompress and structure this latent knowledge. We first detailed a systematic methodology, operationalized by our Socrates agent, to construct a massive, verifiable LCoT Knowledge Base grounded in first principles. We then introduced the Brainstorm Search Engine, a novel tool that performs ``inverse knowledge search" to navigate these reasoning chains, and the Plato agent, which leverages this engine for high-fidelity, low-hallucination creative synthesis. Finally, we demonstrated how these components culminate in SciencePedia, an emergent, deeply interconnected, and scalable scientific encyclopedia.

It is important to position this automated framework primarily as a solution to the critical ``cold-start problem" that plagues traditional, volunteer-driven knowledge projects. While this work demonstrates a scalable method for generating a vast, reliable foundation, we envision the long-term evolution of SciencePedia as a hybrid, collaborative model. Future work will focus on building interfaces that allow the broader scientific community to interact with our agents---enabling experts to contribute, refine content, validate derivations, and correct potential errors (detailed in Appendix \ref{app:mcp}). This synthesis of AI-driven scale and expert-driven accuracy is crucial for ensuring the knowledge base's long-term vitality, comprehensive coverage, and continuous improvement.

While this framework represents a significant step towards a more transparent and interconnected scientific record, we recognize several avenues for future work. The knowledge contained within our system is currently bounded by the training data of the foundational LLMs used in its generation~\cite{knowledge_cutoff1}. To overcome this limitation and push the frontiers of the knowledge base, a promising direction is to apply our rigorous LCoT generation and verification pipeline to other high-quality corpora. Specifically, by systematically digesting advanced textbooks and cutting-edge, peer-reviewed articles, we can continuously update SciencePedia with the latest scientific breakthroughs, ensuring its long-term relevance and accuracy.



Finally, a more ambitious long-term goal is the formalization of the knowledge network. At present, the connections between concepts are implicitly encoded within the natural language of the LCoT reasoning chains~\cite{logical_reasoning1}. A transformative next step would be to apply principles from mathematical logic and automated reasoning to parse and structure these chains into a formal graph. Such an endeavor would move beyond a navigable encyclopedia towards a formalized, machine-readable representation of scientific knowledge, enabling new paradigms of automated discovery and verification.

\backmatter

\bmhead{Acknowledgements}

We thank the DP team for developing the Sciencepedia platform. The authors thank Prof. Gang Su, Prof. Hong Zhao and Prof. Siheng Chen for many invaluable discussions. K. C.
and X. C. are supported by the National Key Research and Development Program of China under Grant
No. 2024YFA1408604 and the National Natural Science Foundation of China (NSFC) under Grants No. 12047503
and No. 12447103. Z. Y. and Y. L. are supported by the NSFC under Grant No. 12247101 and the Natural Science Foundation of Gansu Province No. 22JR5RA389 and No.25JRRA799. Z. Y. also acknowledges the support of the Peng Huanwu Visiting Professor Program, Institute of Theoretical Physics, Chinese Academy of Science. P. Z. is supported by Project 12325501 of the NSFC. Y. D. and S. H. are supported by the NSFC under Grants No. 12275263, the Innovation Program for Quantum Science and Technology under Grant No. 2021ZD0301900, and the Natural Science Foundation of Fujian Province of China under Grant No. 2023J02032. W. E is supported by the NSFC Major Research Project under Grant No. 92270001.

\bibliography{example_paper}
\bibliographystyle{bst/sn-nature}

\begin{appendices}

\section{Uncovering Knowledge's Dark Matter: A Case Study on the Transmon Qubit}
\label{app:transmon}

To make the distinction between our LCoT-based methodology and other approaches concrete, it is instructive to consider a specific, timely example: the transmon qubit. We selected this topic as our case study because its underlying physics---``the discovery of macroscopic quantum mechanical tunnelling and energy quantisation in an electric circuit''---was recognized with the 2025 Nobel Prize in Physics \cite{NobelPhysics2025}.

A simple comparison with a human-curated encyclopedia like Wikipedia is insufficient, as a powerful LLM can often produce a more comprehensive article on a well-established topic. The critical test is to compare the output of our system with that of the \textit{same} LLM using the \textit{same} prompt, but without access to our LCoT knowledge base. This allows us to isolate and demonstrate the unique value added by our framework.

A direct comparison of the four articles reveals a clear hierarchy in knowledge representation, showcasing the spectrum of modern encyclopedic creation. This comparison---spanning traditional human  curation (Wikipedia), LLM-powered curation (Grokipedia), baseline LLM generation, and our LCoT-based synthesis---powerfully underscores the core thesis of this work.

\textbf{Wikipedia: The Compressed Map.} The Wikipedia article is a fact-oriented summary. It is accurate and efficient, defining the transmon, stating its key advantage (charge noise reduction), and providing historical context. It is the epitome of the ``radical compression of reasoning'' we describe in the introduction. It provides the landmarks of the territory but offers no pathways between them. The reader learns \textit{what} a transmon is, but the underlying physics and its broader scientific context---the knowledge ``dark matter''---remain hidden.

\textbf{Grokipedia: The Up-to-Date Map.} The Grokipedia article is a direct adaptation of the Wikipedia entry, preserving the original text while updating technical data such as qubit coherence times and gate fidelities with figures from recent research. This makes it a more current compressed map for experts tracking the state-of-the-art. However, as it is otherwise identical to the Wikipedia article, it does not aim to decompress the underlying reasoning, leaving the ``dark matter'' connections between concepts untouched.

\textbf{Baseline LLM: A Plausible but Generic Narrative.} The baseline LLM article, generated without our knowledge base, represents a significant step up in pedagogical quality. It correctly identifies the core concepts and weaves them into a compelling narrative, starting from the harmonic oscillator and building up to the noise-resilient transmon. However, its description of applications remains generic and high-level (e.g., ``simulating nature'', ``searching for the secrets of the cosmos''). While fluent, it lacks the specific, verifiable, and often surprising connections that characterize deep scientific understanding. It has recreated the \textit{style} of a good explanation but has not synthesized the deep, multi-domain \textit{substance}. This is the ``dark matter blindness'' inherited from its training data.

\textbf{SciencePedia: Uncompressed, Verifiable, and Interconnected.} The SciencePedia article, synthesized from our LCoT knowledge base, demonstrates a fundamentally different quality.
\begin{enumerate}
    \item \textbf{Derivational Depth:} Like the baseline, it follows a logical progression. However, it goes deeper, explaining not just the concepts but the reasoning connecting them. For instance, it details the specific mechanisms of decoherence (Purcell decay, dielectric loss, quasiparticles) and connects the transmon's design ($E_J \gg E_C$) directly to the mitigation of charge noise through an intuitive ``charge reservoir'' analogy. This is a direct consequence of synthesizing from LCoT chains that derive these relationships from first principles.
    \item \textbf{Systematic, Cross-Disciplinary Breadth:} This is the most critical distinction. Where the baseline LLM offers generic applications, the SciencePedia article---drawing on the vast web of connections surfaced by the Brainstorm Search Engine---presents a rich and specific tapestry of the transmon's role across science. It details its use as a ``quantum microscope'' to probe for \textbf{Majorana zero modes}, a tool for the \textbf{ground-state cooling of macroscopic mechanical resonators}, and a miniature laboratory to test the \textbf{Leggett-Garg inequality} and explore \textbf{non-equilibrium quantum thermodynamics}.
\end{enumerate}

These specific, non-obvious applications are not the product of a single generative query but are emergent properties of a knowledge base built on millions of verified reasoning chains. The baseline LLM can write a good story; our system builds the story from a library of verifiable chapters, revealing the profound and often hidden unity of scientific knowledge. This is the essence of search-as-exploration, a resource built not just to find facts, but to discover connections.\\

The articles from all four sources used for direct comparison are presented below. For the Wikipedia entry on the transmon \cite{wikipedia_transmon} and the Grokipedia entry on the same topic \cite{grokipedia_transmon}, only the main text has been retained---ancillary content such as figures, figure captions, and citations has been removed.

\begin{wikipediaarticle}[Wikipedia Transmon Page, Oct. 28, 2025]
\section*{Transmon}

In quantum computing, and more specifically in superconducting quantum computing, a transmon is a type of superconducting charge qubit designed to have reduced sensitivity to charge noise. The transmon was developed by Jens Koch, Terri M. Yu, Jay Gambetta, Andrew Houck, David Schuster, Johannes Majer, Alexandre Blais, Michel Devoret, Steven M. Girvin, and Robert J. Schoelkopf at Yale University and Universit\'{e} de Sherbrooke in 2007. Its name is an abbreviation of the term \textit{transmission line shunted plasma oscillation qubit}; one which consists of a Cooper-pair box ``where the two superconductors are also [capacitively] shunted in order to decrease the sensitivity to charge noise, while maintaining a sufficient anharmonicity for selective qubit control".\\

The transmon achieves its reduced sensitivity to charge noise by significantly increasing the ratio of the Josephson energy to the charging energy. This is accomplished through the use of a large shunting capacitor. The result is energy level spacings that are approximately independent of offset charge. Planar on-chip transmon qubits have $T_1$ coherence times approximately 30 $\mu\mathrm{s}$ to 40 $\mu\mathrm{s}$. Recent work has shown significantly improved $T_1$ times as long as 95 $\mu\mathrm{s}$ by replacing the superconducting transmission line cavity with a three-dimensional superconducting cavity, and by replacing niobium with tantalum in the transmon device, $T_1$ is further improved up to 0.3 $\mathrm{ms}$. These results demonstrate that previous $T_1$ times were not limited by Josephson junction losses. Understanding the fundamental limits on the coherence time in superconducting qubits such as the transmon is an active area of research.\\

\subsection*{Comparison to Cooper-pair box}
The transmon design is similar to the first design of the charge qubitg known as a ``Cooper-pair box"; both are described by the same Hamiltonian, with the only difference being the $E_J/E_C$ ratio. Here $E_J$ is the Josephson energy of the junction, and $E_C$ is the charging energy inversely proportional to the total capacitance of the qubit circuit. Transmons typically have $E_J/E_C \gg 1$ (while $E_J/E_C \lesssim 1$ for typical Cooper-pair-box qubits), which is achieved by shunting the Josephson junction with an additional large capacitor.\\

The benefit of increasing the $E_J/E_C$ ratio is the insensitivity to charge noise---the energy levels become independent of the offset charge $n_g$ across the junction; thus the dephasing time of the qubit is prolonged. The disadvantage is the reduced anharmonicity $\alpha = (E_{21} - E_{10}) / E_{10}$, where $E_{ij}$ is the energy difference between eigenstates $|i\rangle$ and $|j\rangle$. Reduced anharmonicity complicates the device operation as a two level system, e.g. exciting the device from the ground state to the first excited state by a resonant pulse also populates the higher excited state. This complication is overcome by complex microwave pulse design, that takes into account the higher energy levels, and prohibits their excitation by destructive interference. Also, while the variation of $E_{10}$ with respect to $n_g$ tend to decrease exponentially with $E_J/E_C$, the anharmonicity only has a weaker, algebraic dependence on $E_J/E_C$ as $\sim (E_J/E_C)^{-1/2}$. The significant gain in the coherence time outweigh the decrease in the anharmonicity for controlling the states with high fidelity.\\

Measurement, control and coupling of transmons is performed by means of microwave resonators with techniques from circuit quantum electrodynamics also applicable to other superconducting qubits. Coupling to the resonators is done by placing a capacitor between the qubit and the resonator, at a point where the resonator electromagnetic field is greatest. For example, in IBM Quantum Experience devices, the resonators are implemented with ``quarter wave" coplanar waveguides with maximal field at the signal-ground short at the waveguide end; thus every IBM transmon qubit has a long resonator ``tail". The initial proposal included similar transmission line resonators coupled to every transmon, becoming a part of the name. However, charge qubits operated at a similar $E_J/E_C$ regime, coupled to different kinds of microwave cavities are referred to as transmons as well.\\

\subsection*{Transmons as qudits instead of qubits}
Transmons have been explored for use as \textit{d}-dimensional qudits via the additional energy levels that naturally occur above the qubit subspace (the lowest two states). For example, the lowest three levels can be used to make a transmon qutrit; in the early 2020s, researchers have reported realizations of single-qutrit quantum gates on transmonsg as well as two-qutrit entangling gates. Entangling gates on transmons have also been explored theoretically and in simulations for the general case of qudits of arbitrary \textit{d}. 
\end{wikipediaarticle}

\begin{grokipediaarticle}[Grokipedia Transmon Page, Oct. 28, 2025]
\section*{Transmon}
In quantum computing, and more specifically in superconducting quantum computing, a \textbf{transmon} is a type of superconducting charge qubit designed to have reduced sensitivity to charge noise. The transmon was developed by Jens Koch, Terri M. Yu, Jay Gambetta, Andrew Houck, David Schuster, Johannes Majer, Alexandre Blais, Michel Devoret, Steven M. Girvin, and Robert J. Schoelkopf at Yale University and Université de Sherbrooke in 2007. Its name is an abbreviation of the term \textit{transmission line shunted plasma oscillation qubit}; one which consists of a Cooper-pair box ``where the two superconductors are also [capacitively] shunted in order to decrease the sensitivity to charge noise, while maintaining a sufficient anharmonicity for selective qubit control''.\\

The transmon achieves its reduced sensitivity to charge noise by significantly increasing the ratio of the Josephson energy to the charging energy. This is accomplished through the use of a large shunting capacitor. The result is energy-level spacings that are approximately independent of offset charge. As of 2025, optimized transmon qubits achieve $T_1$ relaxation times exceeding $0.5\,\text{ms}$, with records up to $1\,\text{ms}$ in advanced designs using improved materials and fabrication techniques such as high-resistivity substrates and surface encapsulation. These results demonstrate that previous $T_1$ times were not limited by Josephson junction losses. Understanding the fundamental limits on the coherence time in superconducting qubits such as the transmon is an active area of research.

\subsection*{Comparison to Cooper-pair box}
Eigenenergies $E_m$ (first three levels, $m=0,1,2$) of the qubit Hamiltonian as a function of the effective offset charge $n_g$ for different ratios $E_J/E_C$. Energies are given in units of the transition energy $E_{01}$, evaluated at the degeneracy point $n_g=0.5$. The zero point of energy is chosen as the bottom of the $m=0$ level. The charge qubit (small $E_J/E_C$, top) is normally operated at the $n_g=0.5$ ``sweet spot'', where fluctuations in cause less energy shift, and the anharmonicity is maximal. Transmon (large $E_J/E_C$, bottom) energy levels are insensitive to fluctuations, but the anharmonicity is reduced.\\

The transmon design is similar to the first design of the charge qubit known as a ``Cooper-pair box''; both are described by the same Hamiltonian, with the only difference being the $E_J/E_C$ ratio. Here $E_J$ is the Josephson energy of the junction, and $E_C$ is the charging energy inversely proportional to the total capacitance of the qubit circuit. Transmons typically have $E_J/E_C \gg 1$ (while $E_J/E_C \lesssim 1$ for typical Cooper-pair-box qubits), which is achieved by shunting the Josephson junction with an additional large capacitor.\\

The benefit of increasing the $E_J/E_C$ ratio is the insensitivity to charge noise---the energy levels become independent of the offset charge $n_g$ across the junction; thus the dephasing time of the qubit is prolonged. The disadvantage is the reduced anharmonicity $\alpha=(E_{21}-E_{10})/E_{10}$, where $E_{ij}$ is the energy difference between eigenstates $|i\rangle$ and $|j\rangle$. Reduced anharmonicity complicates the device operation as a two-level system; e.g. exciting the device from the ground state to the first excited state by a resonant pulse also populates the higher excited state. This complication is overcome by complex microwave pulse design, that takes into account the higher energy levels, and prohibits their excitation by destructive interference. Also, while the variation of $E_{10}$ with respect to $n_g$ tend to decrease exponentially with $E_J/E_C$, the anharmonicity only has a weaker, algebraic dependence on $E_J/E_C$ as $\sim (E_J/E_C)^{-1/2}$. The significant gain in the coherence time outweigh the decrease in the anharmonicity for controlling the states with high fidelity.\\

Measurement, control and coupling of transmons is performed by means of microwave resonators with techniques from circuit quantum electrodynamics also applicable to other superconducting qubits. Coupling to the resonators is done by placing a capacitor between the qubit and the resonator, at a point where the resonator electromagnetic field is greatest. For example, in IBM Quantum Experience devices, the resonators are implemented with ``quarter wave'' coplanar waveguides with maximal field at the signal-ground short at the waveguide end; thus every IBM transmon qubit has a long resonator ``tail''. The initial proposal included similar transmission line resonators coupled to every transmon, becoming a part of the name. However, charge qubits operated at a similar $E_J/E_C$ regime, coupled to different kinds of microwave cavities are referred to as transmons as well.

\subsection*{Transmons as qudits instead of qubits} 
Transmons have been explored for use as $d$-dimensional qudits via the additional energy levels that naturally occur above the qubit subspace (the lowest two states). For example, the lowest three levels can be used to make a transmon qutrit; in the early 2020s, researchers reported realizations of single-qutrit quantum gates on transmons as well as two-qutrit entangling gates. Entangling gates on transmons have also been explored theoretically and in simulations for the general case of qudits of arbitrary $d$. In 2024, experimental demonstrations included dynamical decoupling protocols for transmon qutrits ($d=3$) and ququarts ($d=4$) to suppress decoherence and crosstalk. Theoretical work has advanced universal pulse schemes for high-fidelity ladder gates in superconducting qudits, targeting fidelities around $10^{-4}$ error for various dimensions.
\end{grokipediaarticle}

\begin{llmbaseline}[Article Generated from an LLM]
\section*{Transmon} 
\subsection*{Key Takeaways}
\begin{itemize}
    \item The transmon qubit's design breakthrough is the addition of a large shunt capacitor to a Josephson junction circuit, which suppresses sensitivity to charge noise and drastically increases quantum coherence time.
    \item By using a non-linear Josephson junction instead of a standard inductor, the transmon becomes an anharmonic oscillator, allowing its lowest two energy states ($|0\rangle$ and $|1\rangle$) to be addressed with microwaves without exciting higher levels.
    \item Transmons are the preferred building blocks for leading superconducting quantum computers, enabling the creation of scalable processors capable of running complex quantum algorithms and simulations.
    \item Beyond computing, transmons act as `artificial atoms' for exploring fundamental physics, serve as transducers to link microwave and optical quantum systems, and function as highly sensitive detectors for fields like radio astronomy.
\end{itemize}

\subsection*{Introduction}
The transmon qubit stands as a triumph of quantum engineering, forming the backbone of many of the world's most powerful quantum computers. While the idea of a quantum bit built from simple electrical components seems straightforward, early designs were plagued by a critical flaw: extreme sensitivity to environmental ``noise", which destroyed their delicate quantum states almost instantly. This roadblock of decoherence threatened to halt the progress of superconducting quantum processors. This article tackles how this challenge was overcome through clever circuit design. In the following chapters, we will first explore the fundamental ``Principles and Mechanisms" of the transmon, tracing its development from a basic quantum harmonic oscillator to a robust, noise-resilient qubit. Subsequently, in ``Applications and Interdisciplinary Connections", we will see how this groundbreaking device is not just a building block for computers but also a versatile tool for simulating nature, exploring fundamental physics, and even searching for the secrets of the cosmos.

\subsection*{Principles and Mechanisms}
Alright, let's peel back the curtain. You might think that to build a quantum computer, you need some fantastically exotic material, some philosopher's stone dug from a meteor. But the heart of many of today's most powerful quantum machines, the \textbf{transmon qubit}, is built from something surprisingly familiar to any electrical engineer: capacitors and inductors. The magic isn't in some rare element, but in how we arrange these simple parts and coax them into obeying the strange and beautiful laws of the quantum world.\\

Our journey to understanding the transmon is really a story in three acts. We start with a perfect, idealized electrical circuit, find its limitations, and then, with a stroke of genius, add just the right ingredients to turn it into a robust quantum bit.

\subsubsection*{The Problem of the Perfect Ladder: The Quantum Harmonic Oscillator}
Imagine a child on a swing. The child swings back and forth, back and forth. The energy of the swing can be anything---a tiny nudge or a massive arc. This is a classic picture of a \textbf{harmonic oscillator}. In the world of electronics, the simplest harmonic oscillator is something called an \textbf{LC circuit}. It consists of an inductor ($L$) and a capacitor ($C$).\\

Think of it this way: the capacitor stores energy in an electric field, like the swing at its highest point (stored potential energy). Then, the charge flows out of the capacitor and through the inductor, which stores energy in a magnetic field, like the swing at its lowest point (maximum kinetic energy). The energy sloshes back and forth between the capacitor and the inductor, oscillating at a natural frequency $\omega_0 = \frac{1}{\sqrt{LC}}$.\\

Now, here's where things get interesting. When we make this circuit very small and very cold, quantum mechanics takes over. Suddenly, the energy of the oscillator can't be just anything. It's \textbf{quantized}. The allowed energies are like the rungs of a perfectly uniform ladder, with each rung separated by the same amount of energy, $\hbar\omega_0$. The energy of the $n$-th rung is given by: 
$$E_n = (n + \frac{1}{2})\hbar\omega_0$$ 
where $n = 0, 1, 2, ...$ and $\hbar$ is the reduced Planck constant. The lowest rung ($n=0$) is the \textbf{ground state}, and the next rung ($n=1$) is the \textbf{first excited state}.\\

It seems we have a natural candidate for a qubit! We can call the ground state $|0\rangle$ and the first excited state $|1\rangle$. To flip the bit from $|0\rangle$ to $|1\rangle$, we just need to give it a little kick of energy---a microwave photon with a precise frequency $\omega_0$ such that $E = \hbar\omega_0$.\\

But there's a terrible flaw in this beautiful plan. Because the rungs of our energy ladder are all equally spaced, a photon with frequency $\omega_0$ will not only drive the transition from $|0\rangle$ to $|1\rangle$, but it will also drive the transition from $|1\rangle$ to $|2\rangle$, from $|2\rangle$ to $|3\rangle$, and so on. If we try to put our qubit into a superposition of $|0\rangle$ and $|1\rangle$, we might accidentally kick it all the way up to $|2\rangle$ or beyond. We can't isolate our two-level system. Our qubit leaks out into a whole ladder of states. We need to break the perfect symmetry of our ladder.

\subsubsection*{The Magic Ingredient: Making the Ladder ``Anharmonic"}
To fix our problem, we need to build an \textbf{anharmonic oscillator}---a system where the energy gaps between successive levels are \textit{not} the same. The gap between $|0\rangle$ and $|1\rangle$ should be different from the gap between $|1\rangle$ and $|2\rangle$. We need to make our ladder crooked.\\

How do we do this in an electrical circuit? We need a non-linear component. A standard inductor is linear; the voltage across it is directly proportional to the rate of change of current. It's too well-behaved. We need something unruly. We need a \textbf{Josephson junction}.\\

A Josephson junction is a marvel of quantum engineering. Imagine taking two pieces of superconducting material and separating them with a sliver of insulating material so thin that pairs of electrons (called \textbf{Cooper pairs}) can ``quantum tunnel" through the barrier without any resistance. This tunneling behavior is profoundly non-linear. Instead of storing energy like a standard inductor ($E \propto \Phi^2$, where $\Phi$ is the magnetic flux), the junction's energy has a cosine dependence:
$$E_J(\delta) = -E_J \cos(\delta)$$
Here, $E_J$ is the \textbf{Josephson energy}, a constant that characterizes the strength of the junction, and $\delta$ is the phase difference of the quantum wavefunction across the junction.\\

When we replace the linear inductor in our LC circuit with a Josephson junction, we create a new circuit---often called a \textbf{Cooper Pair Box}. The cosine nature of the junction's energy warps the potential well of the oscillator. This ``warping" squishes the energy levels together at higher energies. The result is exactly what we wanted: the energy gap between the ground state $|0\rangle$ and the first state $|1\rangle$, let's call the transition frequency $\omega_{01}$, is now \textit{larger} than the gap between $|1\rangle$ and $|2\rangle$ ($\omega_{12}$). The difference, $\alpha = \omega_{01} - \omega_{12}$, is the \textbf{anharmonicity}.\\

Now we have an addressable qubit! We can tune a microwave source to the exact frequency $\omega_{01}$. A pulse at this frequency will drive the qubit between $|0\rangle$ and $|1\rangle$, but it won't have the right energy to promote it from $|1\rangle$ to $|2\rangle$. It's like having a special key that only fits the lock between the first two rungs of the ladder. We have successfully confined our quantum system to the two levels we care about.

\subsubsection*{The Masterstroke: Taming the Noise}
We solved one problem, but created another. The Cooper Pair Box, in its basic form, is fantastically sensitive. It's like a microphone so sensitive it can hear a pin drop from a mile away... but it's sitting in the middle of a rock concert. The ``noise" comes from everywhere. The biggest villain is \textbf{charge noise}---tiny, random fluctuations in the number of stray electrons or charges on the metal island of the qubit itself.\\

These stray charges are a disaster because the energy levels of the Cooper Pair Box depend very strongly on the exact charge present on its central ``island". A single electron hopping on or off can drastically shift the qubit's frequency $\omega_{01}$, destroying any delicate superposition we've worked so hard to create. This loss of quantum information is called \textbf{decoherence}, and it's the ultimate enemy of quantum computation. The lifetime of these early qubits was measured in nanoseconds---hardly long enough to do anything useful.\\

For years, this charge sensitivity seemed like a fundamental roadblock. Then, in 2007, a group of physicists at Yale University had an idea that was both simple and profound. What if we could make the qubit simply \textit{not care} about the stray charges?\\

Their solution was the \textbf{transmon}, short for ``transmission line shunted plasma oscillation qubit". The key insight was to add a very large capacitor in parallel with the Josephson junction. The energy of the qubit is determined by a competition between two scales: the Josephson energy $E_J$, which favors the delocalization of Cooper pairs, and the \textbf{charging energy} $E_C$, which is the energy cost to add a single Cooper pair to the island. In the old Cooper Pair Box, these two energies were roughly comparable ($E_J \approx E_C$).\\

The transmon design makes the charging energy much, much smaller than the Josephson energy by adding the big capacitor. In a typical transmon, the ratio is on the order of $E_J / E_C \approx 50$.\\

What does this do? Think of the capacitor as a giant charge reservoir. The tiny stray charges that were causing so much trouble are now like a single drop of rain falling into a giant lake. Their effect is completely diluted. By making $E_C$ small, the qubit's energy levels become almost independent of the background charge. We've made our qubit nearly deaf to the charge noise that was plaguing it.\\

Of course, there is no free lunch in physics. The price we pay for this noise immunity is a reduction in anharmonicity. But it's a brilliant trade-off. We give up a little bit of addressability to gain a massive improvement in coherence. The lifetime of the qubit jumps from nanoseconds to tens or even hundreds of microseconds---a factor of a thousand or more! That's enough time to perform thousands of quantum gate operations. It was this breakthrough that transformed superconducting qubits from a laboratory curiosity into a viable platform for building scalable quantum computers.\\

So there you have it. The transmon is not so much a single discovery as it is a story of refined thinking. It's a simple harmonic oscillator, made anharmonic with a Josephson junction, and made robust with a large shunt capacitor. It's a testament to the idea that sometimes, the most elegant solutions in physics are about knowing what to ignore.

\subsection*{Applications and Interdisciplinary Connections}
So, we've taken a journey into the heart of the transmon. We’ve seen how this clever little circuit, born from the marriage of a Josephson junction and a capacitor, elegantly sidesteps the noise that plagued its predecessors. We've talked about its energy levels, its anharmonicity, and how we can poke and prod it with microwaves to coax it into quantum states.\\

Now comes the really fun part. We've built a beautiful tool; what can we do with it? You might think its home is solely within the futuristic walls of a quantum computer, and you’d be right---but that’s only the beginning of the story. The transmon is not just a single-purpose gadget. It's more like a master key, one that unlocks doors to entirely new fields of science and engineering. It's a bridge that connects seemingly disparate worlds. Let's walk across a few of these bridges together.

\subsubsection*{The Quantum Architect's Favorite Brick}
First and foremost, the transmon is the star player in the race to build a universal quantum computer. If you think of a classical computer's processor as a city built from billions of tiny transistor switches, then you can think of today's leading quantum processors as small, growing towns built from transmons. These aren't your ordinary on-or-off switches. Each transmon is a qubit, a quantum bit, capable of existing in a superposition of its ground state, $|0\rangle$, and its excited state, $|1\rangle$.\\

Why is it the architect's favorite? Because it hits a sweet spot. Its relatively long coherence time means it can ``remember" its delicate quantum state long enough for us to perform operations on it. Its design allows us to fabricate many of them on a single chip with high precision, opening a believable path toward scaling up from tens of qubits to thousands, and maybe millions. By pulsing microwaves at just the right frequency and duration, we can manipulate individual transmons to perform single-qubit gates, or we can use clever coupling schemes to make two transmons interact, creating the crucial two-qubit gates that are the foundation of complex quantum algorithms.\\

But building a ``digital" quantum computer to run algorithms like Shor's for factoring is only one path. There’s another, perhaps more immediate, application that brings us back to an idea from one of the great minds of physics, Richard Feynman. He once said, ``Nature isn't classical, dammit, and if you want to make a simulation of nature, you'd better make it quantum mechanical!" He was pointing out that our best classical supercomputers struggle mightily to simulate even moderately complex molecules or materials, because the number of variables needed to describe a quantum system explodes exponentially.\\

With a processor made of transmons, we can build a programmable quantum system to simulate \textit{another} quantum system. Want to understand how a new drug molecule will bind to a protein? Configure your transmons to mimic the molecule's electrons and see what they do. Want to design a new material for high-temperature superconductivity? Build a lattice of transmons that behaves like the material's crystal structure and explore its properties directly. It’s a case of using quantum to understand quantum---a beautifully direct and powerful idea.

\subsubsection*{A Playground for Fundamental Physics}
Let's put aside the engineering for a moment and look at the transmon through the eyes of a physicist interested in the fundamental laws of nature. From this perspective, the transmon chip isn't a processor; it's a miniature, controllable universe.\\

When a transmon is placed inside a microwave cavity, it creates something truly remarkable: a near-perfect realization of what physicists call ``cavity quantum electrodynamics", or cavity QED. In a textbook, you learn about a single atom interacting with a single photon of light. This is an astoundingly difficult thing to study with real atoms and a real box of mirrors; the atom flies around, the interaction is weak, and you have very little control.\\

But a transmon in a resonator? Now \textit{that's} a different story. The transmon acts as our ``artificial atom". It stays put. We can tune its energy gap with the flick of a switch (or rather, a magnetic field). The microwave cavity acts as our ``box for light", and we can inject photons one by one. The interaction between our artificial atom and our photons is thousands of times stronger than in typical atomic systems. We have created a QED playground where we can test the fundamental theories of light-matter interaction with an unprecedented level of control. We can watch a single transmon absorb and re-emit a single photon, a quantum dance choreographed entirely by us.\\

This control allows us to explore even deeper questions. The line between the strange quantum world and our familiar classical world is one of the greatest mysteries in physics. A transmon, a macroscopic object you could see under a simple microscope, lives right on this border. We can prepare it in a quintessentially quantum state, like a superposition of two different energy levels---a state analogous to Schr\"{o}dinger's famous cat being both alive and dead. Then, we can literally watch as interactions with the outside world (decoherence) slowly destroy this superposition, forcing the ``cat" to be either alive \textit{or} dead. We are no longer just passive observers of quantum phenomena; we are active participants, able to probe the very fabric of quantum reality.

\subsubsection*{The Universal Translator of the Quantum World}
One of the transmon's most powerful roles is that of an intermediary, a translator between different kinds of quantum systems. The quantum world speaks many languages: the language of microwave photons, the language of optical photons (light), the language of phonons (sound), and the language of electron spins. For quantum technologies to work together, they need a way to communicate. The transmon is shaping up to be the perfect diplomat.\\

Consider a quantum internet. To send quantum information over long distances, we need to use optical photons, because they can travel through fiber-optic cables with very little loss. But our best quantum processors, made of transmons, work with microwave photons. How do you faithfully transfer a quantum state from a transmon to an optical photon without destroying it? You need a quantum transducer. Researchers are building hybrid devices where a transmon is coupled to a tiny, vibrating nano-mechanical object, which in turn is coupled to an optical cavity. The transmon's state, $\alpha|0\rangle + \beta|1\rangle$, gets transferred to the mechanical object, and then from the mechanical object to an optical photon, acting as a translator from the microwave to the optical domain.\\

This ability to ``talk" to mechanical systems is itself a mind-boggling field. By coupling a transmon to a tiny vibrating membrane, the size of a red blood cell, physicists have been able to cool the membrane to its quantum ground state---a state where its motion is essentially frozen to the absolute minimum allowed by the uncertainty principle. Even more wondrous, they can put the entire membrane into a superposition of vibrating and not vibrating. This is quantum mechanics at a scale you can almost imagine seeing, bridging the gap between electrical circuits and tangible, moving objects.

\subsubsection*{The Sentinel at the Quantum Frontier}
Finally, the very property that makes life difficult for a quantum computer designer---the transmon's exquisite sensitivity to its environment---can be turned into its greatest strength. If you want to detect something very, very subtle, you should build a detector that is very, very sensitive.\\

A transmon is so sensitive that it can register the energy from a single microwave photon. By tuning the transmon's energy gap to match the frequency of the photons we want to detect, we can create a near-perfect photon counter. When a photon enters the cavity where the transmon lives, it is absorbed, and the transmon is kicked from its ground state $|0\rangle$ to its excited state $|1\rangle$. We can then easily measure this change of state. Such detectors are invaluable for fields like radio astronomy, searching for faint signals from the distant cosmos.\\

This sensitivity also lets us dream big. Some of the most profound mysteries in cosmology, like the nature of dark matter, might be solved using transmons. One leading candidate for dark matter is a hypothetical particle called the axion. Theory suggests that in the presence of a strong magnetic field, an axion could spontaneously convert into a pair of microwave photons. The signal would be impossibly faint, a whisper in the cosmic noise. But an array of thousands of transmons, all tuned to listen for this specific whisper, could act as a giant ``axion radio". It’s a long shot, but it illustrates the incredible reach of this technology.\\

From building computers to simulating nature, from testing fundamental physics to translating between quantum languages, and from detecting single photons to searching for the secrets of the universe, the transmon is far more than just a component. It is a testament to the beautiful unity of physics, where an idea from condensed matter enables discoveries in cosmology. It is a tool, a toy, and a telescope, all in one, and its story is still just beginning.
\end{llmbaseline}

\begin{sciencepediaarticle}[Article Generated with the aid of our LCoT Knowledge Base]
\section*{Transmon}

\subsection*{Key Takeaways}
\begin{itemize}
    \item A transmon is a superconducting circuit that functions as an artificial atom by leveraging the non-linear potential of a Josephson junction, creating unevenly spaced energy levels (anharmonicity) from which a two-level qubit can be isolated.
    \item The state of a transmon qubit is measured indirectly through a quantum non-demolition technique, where the qubit's state causes a detectable frequency shift in a coupled microwave resonator.
    \item The qubit's performance is fundamentally limited by decoherence, which arises from energy relaxation ($T_1$) due to material defects and quasiparticles, and dephasing ($T_2$) due to environmental charge and magnetic flux noise.
    \item Beyond computation, transmons serve as versatile tools for scientific discovery, acting as sensitive probes for exotic phenomena like Majorana zero modes and as platforms for testing fundamental principles of quantum mechanics and thermodynamics.
\end{itemize}

\subsection*{Introduction}
The quest to build a quantum computer hinges on our ability to create and control artificial atoms, or qubits, that obey the laws of quantum mechanics. Among the most successful candidates is the transmon, a system built not from exotic materials but from a simple superconducting electrical circuit. Its success lies in a clever design that overcomes the fundamental challenge of isolating a two-level quantum system from a circuit that naturally possesses an entire ladder of energy states. This challenge, coupled with the ever-present threat of environmental noise corrupting fragile quantum information, represents the central problem that the transmon's architecture elegantly solves.\\

This article explores the physics and application of the transmon qubit across two main chapters. In ``Principles and Mechanisms", we will dissect the transmon, starting from its basic components---the capacitor and Josephson junction. We will uncover how its quantum nature emerges from the circuit physics and how a crucial property, anharmonicity, allows it to function as a qubit. We will also investigate the methods used to read its state without destroying it and the various noise sources that constantly threaten its coherence. Following this, the chapter on ``Applications and Interdisciplinary Connections" will broaden our view, showcasing the transmon not just as a component but as a powerful tool. We will see how it serves as the workhorse for quantum computers, a sensitive microscope for exploring other quantum systems, and a miniature laboratory for testing the foundations of physics itself.

\subsection*{Principles and Mechanisms}

Imagine you want to build an artificial atom. Not a messy replica with a nucleus and electrons, but something much cleaner, an object that obeys the strange and beautiful laws of quantum mechanics, stripped down to its bare essentials. You want a system with distinct energy levels, where you can put it into one level or another, or even a superposition of both, at will. This is the essence of a qubit, the fundamental building block of a quantum computer. The transmon is perhaps the most successful artificial atom we have built so far, and its secret lies not in exotic materials, but in the clever arrangement of an utterly simple electrical circuit.

\subsubsection*{A Quantum Pendulum in a Circuit}

At its heart, a transmon is just two components: a capacitor and a special element called a Josephson junction. Let's build an analogy. Think of a simple pendulum. It has a mass (which represents inertia) and gravity provides a restoring force that pulls it back to the center. The state of the pendulum can be described by its position (angle) and its momentum.\\

In our transmon circuit, the ``position" is the magnetic flux, $\Phi$, which is related to the voltage across the components. The ``momentum" is the electric charge, $Q$, stored on the capacitor plates. The capacitor, with its capacitance $C$, acts like the mass of the pendulum---it provides the inertia. The more charge you pile onto it, the more energy it costs. The Josephson junction is the quantum equivalent of gravity's restoring force. It consists of two superconductors separated by a whisper-thin insulating barrier. It provides a potential energy that depends on the flux, but not like a simple spring. Its energy is a gentle, rolling landscape described by a cosine function, $V(\Phi) = -E_J \cos(2\pi \Phi/\Phi_0)$, where $E_J$ is the Josephson energy.\\

In classical physics, position and momentum are just numbers you can measure. But in the quantum world, they become operators. Flux and charge are what physicists call \textbf{conjugate variables}, just like position and momentum for a particle. This means they cannot be simultaneously known with perfect precision. Their relationship is cemented by a fundamental law, a commutation relation derived from the circuit's basic physics:

$$
[\hat{\Phi}, \hat{Q}] = i\hbar
$$

This little equation, which we can derive straight from the circuit's description, is the spark of life for our artificial atom. It declares that this simple circuit is not a classical object at all, but a true quantum system. It is this fundamental ``fuzziness" between flux and charge that gives rise to discrete energy levels, just like the quantum nature of an electron bound to a nucleus creates atomic orbitals.

\subsubsection*{The Magic of Anharmonicity: How an Oscillator Becomes a Qubit}

If the Josephson junction's potential well were a perfect parabola (like a simple spring), our circuit would be a ``quantum harmonic oscillator". Its energy levels would be perfectly evenly spaced, like the rungs of an infinite ladder. The energy spacing of this ladder is set by the ``plasma frequency", $\omega_p$, which is determined by the interplay between the circuit's inertia (related to the \textbf{charging energy}, $E_C = (2e)^2/(2C)$) and its restoring force (the \textbf{Josephson energy}, $E_J$). A simple analysis shows this frequency scales as $\omega_p \propto \sqrt{E_J E_C}$. If you send a microwave pulse at this frequency, you don't just excite the system from the ground state ($|0\rangle$) to the first excited state ($|1\rangle$); you drive it up the entire ladder of states. You can't isolate just two levels to make a qubit.\\

This is where the magic happens. The transmon is designed so that the cosine potential is not a perfect parabola. It's slightly flattened at the top, like a real pendulum's swing over a larger angle. We operate in a regime where the Josephson energy is much larger than the charging energy ($E_J \gg E_C$). This makes the quantum fluctuations of the flux small, but large enough to feel that the potential is not perfectly harmonic.\\

The consequence is profound: the energy ladder becomes distorted. The spacing between the rungs is no longer equal. The energy to go from the ground state $|0\rangle$ to the first excited state $|1\rangle$ (let's call its frequency $\omega_{01}$) is now \textit{different} from the energy to go from $|1\rangle$ to the second excited state $|2\rangle$ (frequency $\omega_{12}$). This difference is called the \textbf{anharmonicity}, $\alpha = \omega_{01} - \omega_{12}$.\\

This is the secret to making a qubit! Because $\omega_{01}$ is different from $\omega_{12}$, we can tune a microwave pulse to precisely the frequency $\omega_{01}$. This pulse will selectively drive the transition between $|0\rangle$ and $|1\rangle$, but it will be ``off-resonance" for the transition from $|1\rangle$ to $|2\rangle$, leaving it unaffected. We have successfully isolated a two-level system from the infinite ladder of states. We can now write the Hamiltonian not as a simple oscillator, but as the full, rich system that includes the anharmonicity, which can be solved numerically to find the true, unevenly spaced energy levels. This is our artificial atom, our transmon qubit.

\subsubsection*{Eavesdropping on a Quantum State: The Art of Readout}
So we have a qubit. We can put it in state $|0\rangle$, state $|1\rangle$, or a superposition. But how do we know what state it's in? Looking at it directly would be like trying to see a single photon with your naked eye---the act of observation would instantly destroy the delicate quantum state. We need a more subtle approach, a way to eavesdrop.\\

This is done using another microwave circuit called a \textbf{resonator}. You can think of it as a tiny, high-quality guitar string for microwaves, which vibrates at a very specific frequency, $\omega_r$. We weakly link our transmon qubit to this resonator, typically with a small coupling capacitor, $C_g$.\\

Now, the qubit and the resonator interact. The presence of the qubit slightly alters the properties of the resonator. Here's the key insight, which comes from a phenomenon known as the \textbf{dispersive shift}: the resonant frequency of the resonator \textit{changes} depending on the state of the qubit.\\

Imagine the resonator is a finely tuned violin string. The qubit acts like a tiny clothespin attached to it. If the qubit is in its ground state, $|0\rangle$, the clothespin has a certain ``quantum mass", and the string vibrates at a frequency $\omega_{r,0}$. If the qubit is in its excited state, $|1\rangle$, its quantum properties change, and it's as if the clothespin's mass changes slightly. The string now vibrates at a slightly different frequency, $\omega_{r,1}$.\\

So, to read the qubit's state, we don't poke the qubit. We gently ``pluck" the resonator by sending a weak microwave pulse to it and measuring the frequency of the signal that reflects back. If we measure $\omega_{r,0}$, we know the qubit was in state $|0\rangle$. If we measure $\omega_{r,1}$, we know it was in state $|1\rangle$. This frequency difference, $\delta\omega_{\text{disp}} = \omega_{r,1} - \omega_{r,0}$, is the signal. Crucially, this shift is only possible because of the qubit's anharmonicity, $\alpha$. An ideal harmonic oscillator would produce no state-dependent shift. This method is a beautiful example of a \textbf{quantum non-demolition (QND)} measurement---we learn the state of the qubit without absorbing its energy and destroying it.

\subsubsection*{The Fragile Quantum: A Rogues' Gallery of Noise}
A transmon qubit is a magnificent but fragile object. Its quantum coherence, the very property that makes it powerful, is constantly under attack from the surrounding environment. The lifetime of a quantum state is limited by two main processes: energy relaxation and dephasing.\\

\textbf{Energy Relaxation ($T_1$)}: This is the process of the qubit in state $|1\rangle$ decaying back to the ground state $|0\rangle$, losing its energy to the outside world. It is the fundamental speed limit on how long you can store information. This energy can leak out through several channels:

\begin{itemize}
    \item \textbf{Purcell Decay}: The very resonator we use for readout can act as an antenna, broadcasting the qubit's energy away. The qubit can spontaneously emit a photon into the resonator, which then quickly leaks out into the control lines. The rate of this loss depends strongly on the coupling strength $g$ and how close the qubit's frequency is to the resonator's frequency. Engineers must carefully design their systems to be far ``off-resonance" during computations to minimize this effect.

    \item \textbf{Dielectric Loss}: Nothing is perfect. The materials used to fabricate the qubit---the superconducting metal and the substrate---have microscopic imperfections, often in thin layers on their surfaces. These defects can act like tiny resistors that absorb microwave energy. The amount of energy lost depends on the material's innate \textbf{loss tangent} ($\tan\delta_s$) and the \textbf{participation ratio} ($p_s$), which quantifies how much of the qubit's electric field is concentrated in these lossy regions. The battle for longer-lived qubits is, in large part, a materials science battle against this quantum ``gunk".

    \item \textbf{Quasiparticles}: In a superconductor, electrons are bound together in Cooper pairs. But thermal energy can break these pairs, creating ``quasiparticles"---excited electrons that behave like a tenuous gas. If one of these stray quasiparticles tunnels across the Josephson junction, it can absorb the qubit's energy, causing it to decay. This is why quantum computers must be operated at temperatures near absolute zero, to ``freeze out" as many of these rogue quasiparticles as possible.
\end{itemize}

\textbf{Dephasing ($T_2$)}: Even if the qubit isn't losing energy, it can lose its phase information. Imagine a spinning top. $T_1$ relaxation is like the top falling over. Dephasing is like the top starting to wobble unpredictably. This happens because the qubit's energy gap, $\hbar\omega_{01}$, is not perfectly stable. It jitters in response to environmental noise.

\begin{itemize}
    \item \textbf{Flux Noise}: The transition frequency of many transmons is tuned by an external magnetic flux. Unfortunately, the world is full of fluctuating magnetic fields. This \textbf{flux noise}, often with a characteristic ``$1/f$" power spectrum, causes the qubit's frequency to wander over time. Each time you run an experiment, the qubit's frequency might be slightly different, scrambling the delicate phase relationship that is essential for quantum algorithms.

    \item \textbf{Charge Noise}: Similarly, stray electric charges trapped near the circuit can fluctuate, slightly changing the effective gate charge $n_g$ on the capacitor island. Since the qubit's energy levels have a slight dependence on this charge, this ``charge noise" also contributes to dephasing. The transmon's design with $E_J \gg E_C$ was specifically invented to make the qubit's energy levels exponentially insensitive to this charge noise, a major breakthrough that made these devices practical.
\end{itemize}

Understanding and vanquishing this rogues' gallery of noise is the central challenge in building a fault-tolerant quantum computer. Every improvement in qubit lifetime is a hard-won victory in the ongoing war against decoherence, a testament to our ever-deepening understanding of the beautiful and intricate physics of these remarkable artificial atoms.

\subsection*{Applications and Interdisciplinary Connections}
Having journeyed through the fundamental principles of the transmon qubit, we might be left with the impression of a delicate, artificial atom living in the pristine, isolated world of a physics laboratory. But to leave it there would be like understanding the workings of a transistor without ever seeing a computer. The true wonder of the transmon lies not just in what it \textit{is}, but in what it \textit{does}. Its elegant design, a clever circumvention of nature's noise, has made it a key that unlocks doors into a stunning variety of fields. The transmon is not merely an object of study; it has become a powerful tool, a quantum Swiss Army knife that allows us to build, to probe, and to question the very fabric of reality.\\

In this chapter, we will explore this versatility. We will see how this small superconducting circuit has become the leading protagonist in the quest to build a quantum computer, a sensitive microscope for peering into other exotic quantum worlds, and even a miniature laboratory for testing the foundational principles of physics itself.

\subsubsection*{The Heart of the Quantum Computer}

The most celebrated role of the transmon is as the workhorse of modern quantum computing. If a quantum computer is an orchestra, the transmons are the virtuoso violinists, each capable of playing a beautiful note ($|0\rangle$ or $|1\rangle$) or, more magically, a chord of many notes at once (superposition).\\

Building a powerful quantum computer requires more than just having good qubits. We need them to talk to each other to perform logical operations, or ``gates". And we need them to talk in a precise, controlled way, without gossiping to their neighbors when they're not supposed to. This is where the physics of coupled transmons becomes a story of immense engineering subtlety. The very interaction that allows a ``controlled-NOT" gate between two qubits can also lead to unwanted crosstalk, a conditional frequency shift known as the $ZZ$ interaction. To achieve the high fidelity needed for complex algorithms, we must understand and neutralize even the smallest sources of error. This includes calculating subtle corrections to this interaction that arise from virtual sojourns into higher energy states, such as the $|f\rangle$ state, which we typically try to ignore. Taming these effects is a constant battle, a testament to the level of precision required to make quantum computation a reality.\\

Beyond computation within a single processor, the future of quantum computing likely lies in modular architectures---networks of smaller quantum processors linked together. How do you move a fragile quantum state from one module to another? Here again, the transmon shines, this time as a quantum mediator. By cleverly tuning its frequency, a transmon can act as a switchable ``coupler", grabbing a single photon from one microwave cavity and faithfully releasing it into another. This allows for the high-fidelity transfer of quantum states, a fundamental building block for a ``quantum internet".\\

Of course, the nemesis of any quantum computation is noise from the environment, which corrupts the delicate quantum states. This is not just an abstract idea; it has concrete consequences for the algorithms we wish to run. Consider a cornerstone algorithm like Shor's order-finding routine. Its performance is directly tied to a subroutine called Quantum Phase Estimation. When implemented on a real transmon processor, the unavoidable low-frequency magnetic flux noise (so-called $1/f$ noise) introduces small, correlated phase errors in the quantum gates. By modeling how this specific type of physical noise propagates through the algorithm, we can directly predict the decline in the final fidelity, linking the messy reality of the hardware to the abstract performance of the software. This connection is where the physicist's understanding of noise meets the computer scientist's demand for performance.\\

Even the protocols designed to fight noise---quantum error correction codes---create their own unique signatures. In a surface code, for instance, data qubits are periodically ``checked" by neighboring measurement qubits. These check-up operations perturb the energy levels of the data transmon, causing its resonance frequency to shift back and forth. A transmon that is being constantly driven will see its famous three-peaked Mollow triplet fluorescence spectrum ``breathe" in time with the error correction cycle, with the power shifting between the central peak and the sidebands as the probing sequence unfolds. The light emitted by the qubit thus becomes a direct reporter on the error correction code being executed upon it.

\subsubsection*{A Quantum Microscope for Exotic Worlds}
The same sensitivity that makes the transmon susceptible to noise also makes it an extraordinarily sensitive detector. By coupling a transmon to another quantum system, we can use the transmon as a ``quantum microscope", measuring tiny shifts in its frequency to learn about the system it is touching.\\

One of the most thrilling frontiers is the search for Majorana zero modes, exotic particle-like excitations predicted to exist in topological superconductors. These Majoranas are candidates for building inherently fault-tolerant topological qubits. But how do you confirm you've created such an elusive entity? You can couple the topological island hosting the Majoranas to a transmon. The state of the Majorana qubit---whether it is storing a logical $|0\rangle$ or $|1\rangle$---will slightly alter the electrostatic environment of the transmon, causing a minuscule but measurable shift in its frequency. By carefully measuring this frequency shift, the transmon effectively ``reads out" the state of the mystical Majorana qubit, confirming its existence and properties.\\

This principle extends to creating ``hybrid quantum systems" that combine the best of different worlds. While transmons offer fast gate operations, other quantum systems, like the spin of a Nitrogen-Vacancy (NV) center in a diamond, boast exceptionally long coherence times---they can ``remember" their quantum state for much longer. Can we have both? By placing a transmon and an NV center in the same microwave cavity, they can talk to each other by exchanging virtual photons through the cavity. This establishes an effective interaction between them, allowing the transmon to act as a fast processor and the NV center to act as a robust quantum memory.\\

The transmon's ability to control other quantum systems extends even into the realm of macroscopic motion. It is a remarkable achievement of modern physics that we can cool a tiny mechanical object---a vibrating membrane or beam comprising billions of atoms---so close to absolute zero that it enters its quantum ground state of motion. The transmon plays a starring role in this process of ``ground-state cooling". By coupling a transmon to the mechanical resonator, we can use the transmon as a conduit to pump entropy out of the mechanical system. Both the transmon and an associated microwave cavity can be used simultaneously as ``refrigerators", with their own decay channels providing pathways to dump the thermal energy of the resonator into a cold bath, achieving even lower final phonon numbers. Here, the transmon helps bridge the gap between our quantum world and the classical world of tangible, moving objects.

\subsubsection*{A Laboratory for Fundamental Physics}
Finally, the precise control we have over transmons makes them perfect miniature laboratories for exploring the foundational questions of quantum mechanics and thermodynamics. They are not just tools for technology, but tools for pure discovery.\\

Quantum mechanics famously clashes with our classical intuition. One such clash is formalized by the Leggett-Garg inequality, which tests the principle of ``macrorealism"---the common-sense idea that a macroscopic object has a definite state at all times, whether we are looking at it or not. Can a transmon, an object we might consider ``macroscopic" compared to an electron, be said to be oscillating between its states even when we don't measure it? By performing a series of measurements on a transmon and analyzing the time correlations between the outcomes, we can find that they violate the bound set by macrorealism. The transmon, through its coherent quantum evolution, demonstrates that our classical intuition fails, even for a system made of trillions of atoms.\\

The energy levels of a quantum system are not static. They can be pushed and pulled by external fields, creating a new, ``dressed" reality. When two transmons are coupled and one is hit with a strong microwave drive, the energy levels of the whole system rearrange. Probing the other qubit reveals that its transition is no longer a single sharp line, but has split into two, a phenomenon known as the Autler-Townes effect. The size of this splitting is a direct measure of the interplay between the intrinsic coupling of the qubits and the external drive, providing a beautiful spectroscopic window into the dressed-state picture of quantum optics.\\

Even the laws of thermodynamics take on a new, richer meaning in the quantum realm. Consider a single transmon, continuously driven by a microwave field and simultaneously losing energy to its environment. This system reaches a non-equilibrium steady state, a dynamic balance where the power injected by the drive is exactly balanced by the power dissipated. This dissipated power, which can be calculated from the system's dynamics, is directly related to the rate of entropy production. The transmon becomes a controllable model system for investigating the frontier of non-equilibrium quantum thermodynamics, allowing us to explore the fundamental nature of heat, work, and information flow at the single-quantum level.\\

From the heart of a future computer to a probe of ghost-like particles and a testbed for the nature of reality, the transmon has proven to be an astonishingly rich and unifying concept. Its story is a profound illustration of how a deep understanding of one corner of physics can ripple outwards, providing the tools and insights to revolutionize countless others.
\end{sciencepediaarticle}

\section{MCP tools}
\label{app:mcp}

This appendix provides detailed specifications for the Model Context Protocol (MCP) tools utilized in our framework. These tools serve as standardized interfaces for interacting with the core functionalities provided by our LLM agents, enabling modularity and interoperability.

\subsection{Article writer  (Plato Agent Functionality)}

The \verb|generate_article| tool encapsulates the functionality of the Plato agent, designed to synthesize scientific articles based on specified criteria. 

\paragraph{Input Parameters:}
\begin{itemize}
    \item \verb|topic| (string, mandatory): Defines the central subject of the article to be generated (e.g., \verb|"Quantum Tunneling"|).
    \item \verb|language| (string, mandatory): Specifies the desired output language for the article (e.g., \verb|"en-US"|, \verb|"zh-CN"|).
    \item \verb|style_guide| (string, optional): Provides specific instructions or constraints on the writing style, tone, or formatting (e.g., \verb|"Write in the style of Feynman Lectures"|). If omitted, a default pedagogical style is applied.
    \item \verb|model_name| (string, optional): Allows specifying a particular underlying LLM for generation, potentially enabling experimentation with different model capabilities.
\end{itemize}

\paragraph{Return Object (\texttt{ArticleContent}):} The tool returns a structured object containing the generated article and associated metadata.
\begin{itemize}
    \item \verb|topic| (string): Mirrors the input topic for context.
    \item \verb|style_guide| (string): The style guide that was applied during generation (either user-provided or the default).
    \item \verb|language| (string): The language of the generated \verb|main_content|.
    \item \verb|model_name| (string): The specific LLM used for this generation task.
    \item \verb|main_content| (string): The full text body of the synthesized encyclopedia article.
\end{itemize}





\subsection{Problem Generation and Solving (Socrates Agent Functionality)}

The \verb|generate_problems| and \verb|solve_problems| tools collectively implement the core functionality of the Socrates agent described in the main text. These tools support the Socratic method used in constructing the LCoT knowledge base by generating relevant questions (problems) and their corresponding detailed solutions based on first principles, facilitating the generation and cross-validation of LCoT entries.

\paragraph{Problem Generator (\texttt{generate\_problems}):} This tool generates a set of problems within a specified scientific domain.
\subparagraph{Input Parameters:}
\begin{itemize}
    \item \verb|subject| (string, mandatory): Specifies the broad academic discipline (e.g., \verb|"computational physics"|).
    \item \verb|field| (string, mandatory): Narrows down the specific subdomain within the subject (e.g., \verb|"quantum tunneling"|).
    \item \verb|count| (integer, mandatory): The target approximate number of distinct problems to generate (not a strict limit; the actual number may be lower).
    \item \verb|education_level| (string, optional, default: \verb|"advanced_undergraduate"|): Specifies the target educational context for the problems.
\end{itemize}
\subparagraph{Return Object (List of \texttt{Problem}):} Returns a list, where each item is a \verb|Problem| object representing a generated question and its initial solution attempt by the generator itself.
\begin{itemize}
    \item \verb|task_id| (integer): A unique identifier assigned to each generated problem.
    \item \verb|problem| (string): The textual statement of the problem or question.
    \item \verb|answer_type| (string): Indicates the expected format of the answer (e.g., \verb|"multiple_choice"|, \verb|"calculation"|, \verb|"code"|).
    \item \verb|solution| (string): The detailed step-by-step solution derived by the \textbf{problem generator} agent.
    \item \verb|answer| (string): The concise final answer extracted from the \verb|solution| by the \textbf{problem generator}.
\end{itemize}

\paragraph{Problem Solver (\texttt{solve\_problems}):} This tool takes existing problems and generates independent solutions, typically used for cross-validation against the solutions from \verb|generate_problems|.
\subparagraph{Input Parameters:}
\begin{itemize}
    \item \verb|subject| (string, mandatory): Same as in \verb|generate_problems|.
    \item \verb|field| (string, mandatory): Same as in \verb|generate_problems|.
    \item \verb|problems| (list of \verb|Problem| objects, mandatory): A list of problems to be solved. Requires at least \verb|task_id|, \verb|problem|, and \verb|answer_type| attributes for each \verb|Problem| object. Can directly accept results from \verb|generate_problems|.
\end{itemize}
\subparagraph{Return Object (List of \texttt{Problem}):} Returns the input list of \verb|Problem| objects, but now populated with the \verb|solution| and \verb|answer| generated by the \textbf{solver} agent. This allows direct comparison with the generator's output for verification.

\subsection{Code Execution (Runtime Utilities)}
The following tools provide a lightweight, sandboxed runtime for executing short code snippets and computing scores at scale. They are typically used to (i) validate algorithmic steps produced in problem solutions for cross–validation within the LCoT pipeline, (ii) run quick numerical experiments, and (iii) batch–score candidate solutions.

\paragraph{List Supported Languages(\texttt{list\_supported\_languages}):} This tool enumerates the programming languages currently available in the MCP server runtime.
\subparagraph{Return Object (JSON string):} A JSON string containing the list of supported languages.

\paragraph{Simgle Code Execute(\texttt{execute\_code}):} Executes a single code snippet in the specified language.
\subparagraph{Input Parameters:}
\begin{itemize}
    \item \verb|language| (string, mandatory): Programming language (e.g., \verb|"python"|, \verb|"c"|, \verb|"julia"|, \verb|"lean"|).
    \item \verb|code| (string, mandatory): Source code to execute.
    \item \verb|timeout| (integer): Timeout in seconds (default: 10.0).
\end{itemize}
\subparagraph{Return Object (JSON string):} A JSON string with execution result.

\paragraph{Parallel Code Execution(\texttt{execute\_codes\_parallel}):} Executes multiple code snippets in parallel.
\subparagraph{Input Parameters:}
\begin{itemize}
    \item \verb|language| (string, mandatory): Programming language (e.g., \verb|"python"|, \verb|"c"|, \verb|"julia"|, \verb|"lean"|).
    \item \verb|code_list| (list of \verb|code| string, mandatory): List of code strings to execute.
    \item \verb|timeout| (integer): Timeout in seconds for each code execution (default: 10.0).
\end{itemize}
\subparagraph{Return Object (JSON string):} A JSON string containing the list of execution results.

\paragraph{Compute Score Parallel(\texttt{compute\_score\_parallel}):} Executes the \verb|compute_score| function in parallel for multiple solutions.
\subparagraph{Input Parameters:}
\begin{itemize}
    \item \verb|data_source| (string, mandatory): Data source identifier (e.g., \verb|"theoretical_physics"|)
    \item \verb|solution_list| (list of \verb|solution| string, mandatory): List of solution strings to score
    \item \verb|ground_truth_list| (list of \verb|ground truth answer| string, mandatory): List of ground truth strings for comparison
    \item \verb|extra_info_list| (list of \verb|extra info| string, mandatory): Optional list of extra info dictionaries for each solution
    \item \verb|timeout| (integer): Timeout in seconds for each score computation (default: 30.0).
\end{itemize}
\subparagraph{Return Object (JSON string):} A JSON string containing the list of score results including execution time.

\subsection{Usage and Integration}

Several methods facilitate the use and integration of these MCP tools into development workflows and AI agent applications:
\begin{itemize}
    \item \textbf{MCP Inspector:} A graphical tool (\url{https://github.com/modelcontextprotocol/inspector}) for developers to directly interact with and debug MCP tools and servers.
    \item \textbf{Python SDK:} Provides programmatic access (\url{https://github.com/modelcontextprotocol/python-sdk}) for integrating MCP tool calls within AI agent logic, compatible with frameworks like Google ADK (\url{https://google.github.io/adk-docs/}) and LangGraph (\url{https://www.langchain.com/langgraph}).
    \item \textbf{Asynchronous Execution:} For handling potentially long-running generation or solving tasks without blocking, the \verb|bohr-agent-sdk| (\url{https://github.com/dptech-corp/bohr-agent-sdk}) can be used to invoke these tools asynchronously.
\end{itemize}

\end{appendices}



\end{document}